\documentclass[pdflatex,sn-mathphys-num]{sn-jnl}

\usepackage{graphicx}%
\usepackage{multirow}%
\usepackage{amsmath,amssymb,amsfonts}%
\usepackage{amsthm}%
\usepackage{mathrsfs}%
\usepackage[title]{appendix}%
\usepackage{xcolor}%
\usepackage{textcomp}%
\usepackage{manyfoot}%
\usepackage{booktabs}%
\usepackage{algorithm}%
\usepackage{algorithmicx}%
\usepackage{algpseudocode}%
\usepackage{listings}%
\usepackage{tabularx}%
\usepackage{newfloat}%
\usepackage{caption}%

\usepackage{enumitem}
\usepackage{subcaption}

\usepackage{titletoc} 
\newcommand{\appendixtoc}{
  \clearpage
  \startcontents[appendix]
  \section*{Appendices}
  \printcontents[appendix]{}{1}{\setcounter{tocdepth}{3}}
}

\DeclareFloatingEnvironment[
    fileext=lof,
    listname={List of Extended Data Figures},
    name=Extended Data Figure
]{extfigure}

\theoremstyle{thmstyleone}%
\theoremstyle{thmstyletwo}%

\theoremstyle{thmstylethree}%

\begin{document}

\title[Assessing the potential of deep learning for protein-ligand docking]{
    \centering
    Assessing the potential of deep learning\\
    for protein-ligand docking
}


\author*[1]{\fnm{Alex} \sur{Morehead}}\email{acmwhb@lbl.gov}
\author[1]{\fnm{Nabin} \sur{Giri}}\email{ngiri@lbl.gov}
\author[2]{\fnm{Jian} \sur{Liu}}\email{jl4mc@missouri.edu}
\author[2]{\fnm{Pawan} \sur{Neupane}}\email{pngkg@missouri.edu}
\author*[2]{\fnm{Jianlin} \sur{Cheng}}\email{chengji@missouri.edu}

\affil*[1]{\orgname{Lawrence Berkeley National Laboratory}, \orgaddress{\city{Berkeley}, \state{California}, \country{USA}}}

\affil[2]{\orgdiv{Electrical Engineering \& Computer Science, NextGen Precision Health}, \orgname{University of Missouri}, \orgaddress{\city{Columbia}, \state{Missouri}, \country{USA}}}


\abstract{
The effects of ligand binding on protein structures and their \textit{in vivo} functions carry numerous implications for modern biomedical research and biotechnology development efforts such as drug discovery. Although several deep learning (DL) methods and benchmarks designed for protein-ligand docking have recently been introduced, to date no prior works have systematically studied the behavior of the latest docking and structure prediction methods within the \textit{broadly applicable} context of (1) using predicted (apo) protein structures for docking (e.g., for applicability to new proteins); (2) binding multiple (cofactor) ligands concurrently to a given target protein (e.g., for enzyme design); and (3) having no prior knowledge of binding pockets (e.g., for generalization to unknown pockets). To enable a deeper understanding of docking methods' real-world utility, we introduce \textsc{PoseBench}, the first comprehensive benchmark for \textit{broadly applicable} protein-ligand docking. \textsc{PoseBench} enables researchers to rigorously and systematically evaluate DL methods for apo-to-holo protein-ligand docking and protein-ligand structure prediction using \textit{both} primary ligand and multi-ligand benchmark datasets, the latter of which we introduce for the first time to the DL community. Empirically, using \textsc{PoseBench}, we find that (1) DL co-folding methods generally outperform comparable conventional and DL docking baseline algorithms, yet popular methods such as AlphaFold 3 are still challenged by prediction targets with novel protein-ligand binding poses; (2) certain DL co-folding methods are highly sensitive to their input multiple sequence alignments, while others are not; and (3) DL methods struggle to strike a balance between structural accuracy and chemical specificity when predicting novel or multi-ligand protein targets. Code, data, tutorials, and benchmark results are available at \url{https://github.com/BioinfoMachineLearning/PoseBench}.}


\maketitle

\section{Introduction}\label{section:introduction}

The field of drug discovery has long been challenged with a critical task: determining the structure of ligand molecules in complex with proteins and other key biomolecules \citep{warren2012essential}. As accurately identifying such complex structures (in particular multi-ligand structures) can yield advanced insights into the binding dynamics and functional characteristics (and thereby, the medicinal potential) of numerous protein complexes \textit{in vivo}, in recent years, significant resources have been spent developing new experimental and computational techniques for protein-ligand structure determination \citep{du2016insights}. Over the last decade, machine learning (ML) methods for structure prediction have become indispensable components of modern structure determination at scale, with AlphaFold 2 for protein structure prediction being a hallmark example \citep{jumper2021highly, abriata2024nobel}.

As the field has gradually begun to investigate whether proteins in complex with other types of molecules can faithfully be modeled with ML (and particularly deep learning (DL)) techniques \citep{dhakal2022artificial, harris2023benchmarking, krishna2024generalized}, several new works in this direction have suggested the promising potential of such approaches to protein-ligand structure determination \citep{corso2022diffdock, lu2024dynamicbind, qiao2024state, abramson2024accurate}. Nonetheless, it remains to be shown the extent to which the latest of such (docking and co-folding-based) DL methods can adequately generalize to the context of binding novel or uncommon protein-ligand interaction (PLI) pockets and multiple interacting ligand molecules (e.g., which can alter the chemical functions of various enzymes) as well as whether such methods can faithfully model amino acid-specific types of PLIs natively found in crystallized biomolecular structures.

\begin{figure}
  \centering
  \includegraphics[width=\linewidth]{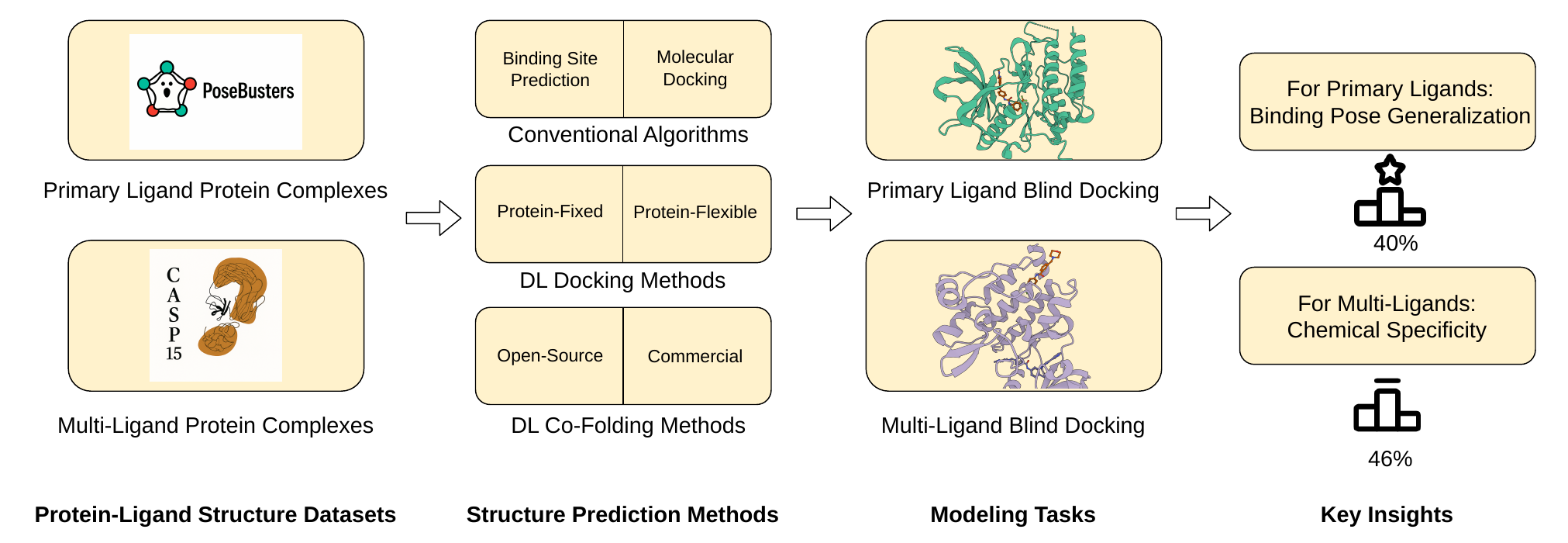}
  \caption{Overview of \textsc{PoseBench}, our comprehensive benchmark for \textit{broadly applicable} DL modeling of primary and multi-ligand protein complex structures. Baseline algorithms within the benchmark include a range of the latest DL docking and co-folding methods, both open-source and commercially restrictive, as well as conventional algorithms for docking. Key observations derived using \textsc{PoseBench} include the discontinuity between structure and interaction modeling performance for novel or uncommon prediction targets and the heavy reliance of key DL co-folding methods on MSA-based input features to achieve high structural accuracy.}
  \label{figure:posebench}
\end{figure}

To bridge this knowledge gap, our contributions in this work are as follows: \\
\begin{itemize}
    \item We introduce the first unified benchmark for protein-ligand docking and structure prediction that evaluates the performance of several recent DL-based baseline methods (DiffDock-L, DynamicBind, NeuralPLexer, RoseTTAFold-All-Atom, Chai-1, Boltz-1, and AlphaFold 3) as well as conventional algorithms (P2Rank + AutoDock Vina) for primary and \textit{multi}-ligand docking, which suggests that DL co-folding methods generally outperform conventional algorithms yet remain challenged by novel or uncommon prediction targets.
    \item In contrast to several recent works using crystal protein structures for protein-ligand docking \citep{buttenschoen2024posebusters, corso2024deep}, the docking benchmark results we present in this work are all within the context of \textit{standardized} input multiple sequence alignments (MSAs) and high accuracy \textit{apo-like} (i.e., AlphaFold 3-predicted) protein structures (see Supplementary Appendix \ref{section:appendix_documentation_for_datasets}) without specifying known binding pockets, which notably enhances the broad applicability of this study's findings.
    \item Our newly proposed benchmark, \textsc{PoseBench}, enables specific insights into necessary areas of future work for accurate and generalizable biomolecular structure prediction, including that DL methods struggle to balance faithful modeling of native PLI fingerprints (PLIFs) with structural accuracy during pose prediction and that some DL co-folding methods (AlphaFold 3) are more dependent than others (Boltz-1, Chai-1) on the availability of input MSAs.
    \item Our benchmark results also highlight the importance of including challenging (out-of-distribution) datasets when evaluating future DL methods while measuring their ability to recapitulate amino acid-specific PLIFs with an appropriate new metric that we introduce in this work.
\end{itemize}

\section{Related work}\label{section:related_work}

\textbf{Structure prediction of PLI complexes.} The field of DL-driven protein-ligand structure determination was largely sparked with the development of geometric deep learning methods such as EquiBind \citep{stark2022equibind} and TANKBind \citep{lu2022tankbind} for direct (i.e., regression-based) prediction of bound ligand structures in protein complexes. Notably, these predictive methods could estimate localized ligand structures in complex with multiple protein chains as well as the associated complexes' binding affinities. However, in addition to their limited predictive accuracy, they have more recently been found to frequently produce steric clashes between protein and ligand atoms, notably hindering their widespread adoption in modern drug discovery pipelines.

\textbf{Protein-ligand structure prediction and docking.} Shortly following the first wave of predictive methods for protein-ligand structure determination, DL methods such as DiffDock \citep{corso2022diffdock} demonstrated the utility of a new approach to this problem by reframing protein-ligand docking as a generative modeling task, whereby multiple ligand conformations can be generated for a particular protein target and rank-ordered using a predicted confidence score \citep{yim2024diffusion}. This approach has inspired many follow-up works offering alternative formulations of this generative approach to the problem \citep{zhang2023efficient, masters2023fusiondock, plainer2023diffdock, guo2023diffdock, pei2024fabind, lu2024dynamicbind, zhu2024diffbindfr, cao2024surfdock, qiao2024state, huang2024re, minan2024geodirdock, corso2024deep, krishna2024generalized, bryant2024structure, starkharmonic, morehead2024flowdock, corso2024flexible, abramson2024accurate, chai2024chai, wohlwend2024boltz, qiao2024neuralplexer3, passaro2025boltz}, with some of such follow-up works also being capable of accurately modeling protein flexibility upon ligand binding or predicting binding affinities to a high degree of accuracy.

\textbf{Benchmarking efforts for protein-ligand complexes.} In response to the large number of new methods that have been developed for protein-ligand structure prediction, recent works have introduced several new datasets and metrics with which to evaluate newly developed methods \citep{vskrinjar2025have}, with some of such benchmarking efforts focusing on modeling single-ligand protein interactions \citep{yu2023deep, buttenschoen2024posebusters, durairaj2024plinder, errington2024assessing, jain2024deep, sharon2024go, hu2024opendock} and others specializing in the assessment of multi-ligand protein interactions \citep{robin2023assessment}. One of the motivations for introducing \textsc{PoseBench} in this work is to bridge this gap by systematically assessing a selection of the latest (pocket-blind) structure prediction methods within both interaction regimes, using unbound (apo) protein structures with docking methods and challenging DL co-folding methods to predict full bioassemblies from primary sequences. Our benchmarking results in the following Section \ref{section:results_and_discussion} demonstrate the relevance and utility of this comprehensive new evaluation suite for the future of protein-ligand modeling.

\section{Results and discussion}\label{section:results_and_discussion}

\begin{figure}
  \centering
  \includegraphics[width=\linewidth]{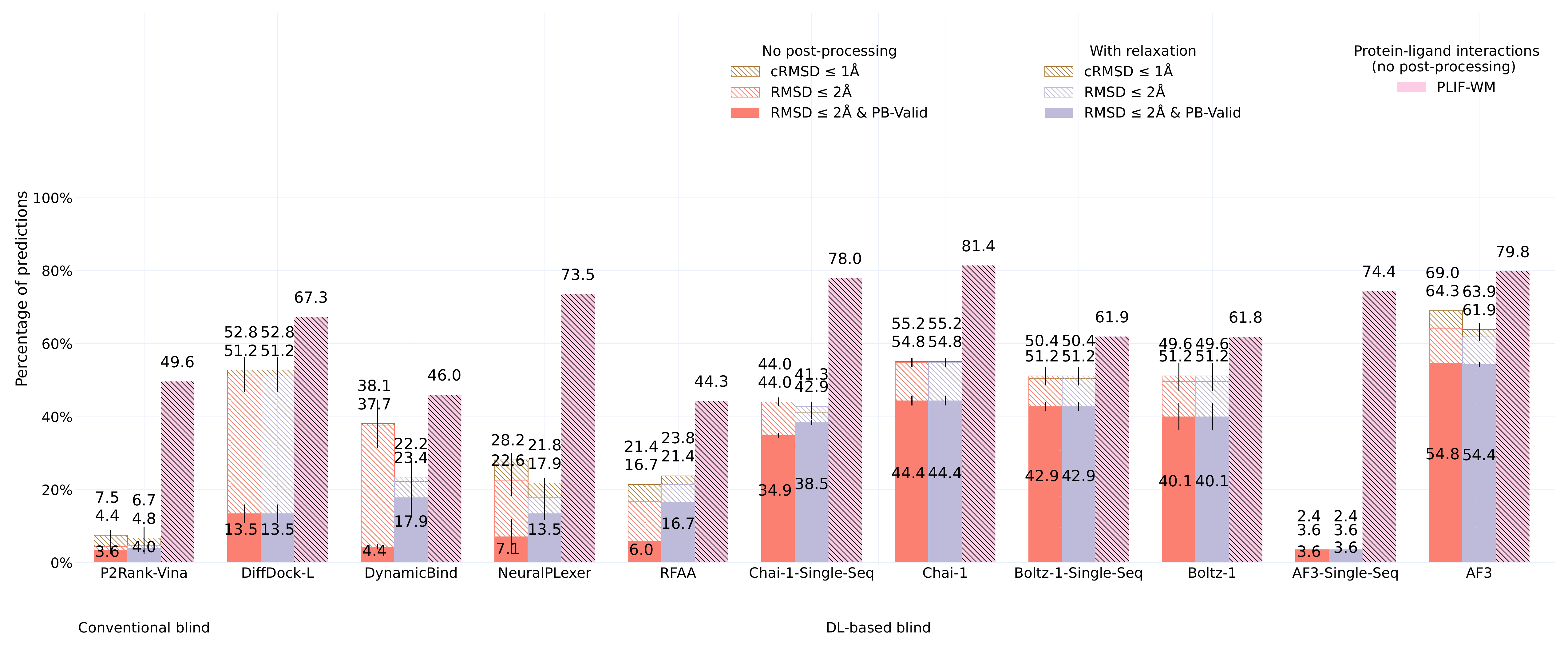}
  \caption{Astex Diverse primary ligand docking success rates (n=85 protein-ligand complexes). Data are presented as mean values +/- standard deviations over three independent predictions for each complex.}
  \label{figure:astex_diverse_results}
\end{figure}

In this section, we present \textsc{PoseBench}'s results for primary and multi-ligand protein-ligand docking and structure prediction and discuss their implications for future work, as succinctly illustrated in Figure \ref{figure:posebench}. Note that across all experiments, for generative methods, we report their performance metrics in terms of the mean and standard deviation across \textit{three} independent runs of each method to gain insights into their inter-run stability and consistency. Key metrics include a method's percentage of structurally accurate ligand pose predictions with a (heavy atom centroid) root mean square deviation (RMSD) less than 2 (1) \AA\ (i.e., (c)RMSD $\leq$ 2 (1) \AA); its percentage of structurally accurate pose predictions that are also chemically valid according to the PoseBusters software suite (i.e., RMSD $\leq$ 2 \AA\ \& PB-Valid), which can be affected by the post-hoc application of structural relaxation driven by computationally expensive molecular dynamics (MD) simulations \citep{eastman2010openmm} (i.e., with relaxation); and our newly proposed Wasserstein matching score of its amino acid-specific predicted PLIFs (PLIF-WM). We formally define these metrics in Section \ref{section:metrics}. For interested readers, in Supplementary Appendix \ref{section:appendix_compute_resources}, we report the average runtime and memory usage of each baseline method to determine which methods are the most efficient for real-world structure-based applications, and in Supplementary Appendix \ref{section:appendix_additional_results} we present supplementary results.

\subsection{Astex Diverse results}\label{section:astex_diverse_results}

Containing PLI structures deposited in the RCSB Protein Data Bank (PDB) \citep{bank1971protein} up until 2007, most of the well-known Astex Diverse dataset's structures \citep{hartshorn2007diverse} are present in the training data of each baseline DL method, and accordingly benchmarking results for this dataset (n=85 protein-ligand complexes), shown in Figure \ref{figure:astex_diverse_results}, indicate that DL co-folding and DL docking methods achieve higher structural and chemical accuracy rates (RMSD $\leq$ 2 \AA\ \& PB-Valid) than the conventional docking baseline AutoDock Vina combined with P2Rank for PLI binding site prediction to facilitate blind molecular docking. Importantly, most DL co-folding methods can reliably identify the correct PLI binding pocket at least 50\% of the time, where DL co-folding methods Chai-1 \citep{chai2024chai}, Boltz-1 \citep{wohlwend2024boltz}, and AlphaFold 3 (AF3) \citep{abramson2024accurate} achieve a reasonable balance between their rates of structural and chemical accuracy and chemical specificity (PLIF-WM), with the single-sequence (i.e., MSA-ablated) version of AF3 being a notable exception. These results suggest that DL co-folding methods have learned the most comprehensive representations of this dataset's input sequences, while at the same time only the DL co-folding method AF3 is significantly impacted by an unavailability of diverse input MSAs. One potential explanation for this phenomenon is that AF3's training relied upon the availability of features derived from protein \textit{and} RNA input MSAs, which may have imbued the model with strongly MSA-\textit{dependent} representations for biomolecular structure prediction.

\subsection{DockGen-E results}\label{section:dockgen_results}

\begin{figure}
  \centering
  \includegraphics[width=\linewidth]{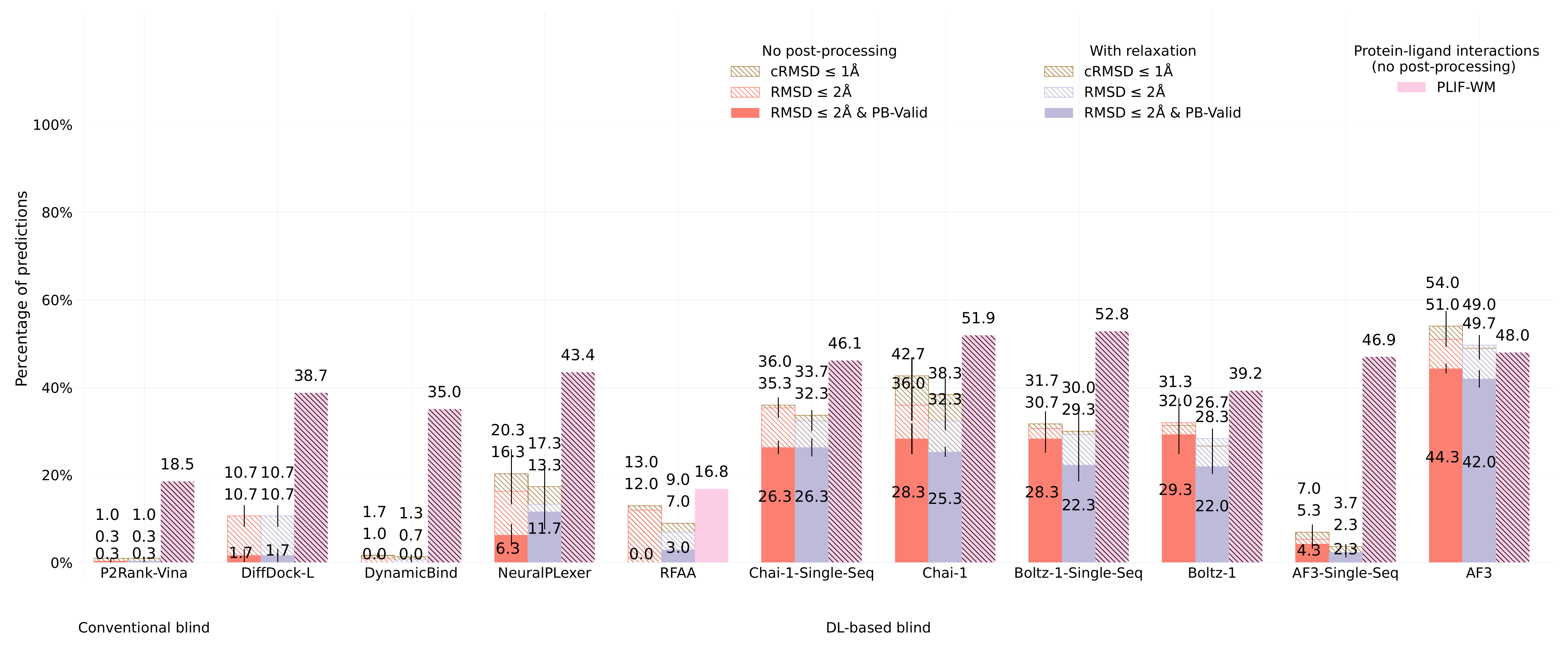}
  \caption{DockGen-E primary ligand docking success rates (n=122 protein-ligand complexes). Data are presented as mean values +/- standard deviations over three independent predictions for each complex.}
  \label{figure:dockgen_e_results}
\end{figure}

As visualized in Figure \ref{figure:dockgen_e_results}, results with our new DockGen-E dataset of biologically relevant PLI complexes deposited in the PDB up to 2019 (n=122 protein-ligand complexes) demonstrate that only the latest DL co-folding methods can locate a sizable fraction of structurally accurate PLI binding poses represented in this dataset. As such methods may have previously seen these PLI structures in their respective training data, it is surprising that even the latest AF3 model fails to identify a structurally and chemically accurate pose for more than 50\% of the dataset's complexes. Further, for Chai-1 and Boltz-1, their single-sequence variants achieve performance comparable to their MSA-based versions, which indicates that for these methods MSA features may not provide enough signal to reliably locate uncommon PLI binding pockets. The overall lower range of PLIF-WM values achieved by each method for this dataset further suggests the increased chemical modeling difficulty of this dataset's complexes compared to those presented by the Astex Diverse dataset. A potential source of these difficulties is that each of this dataset's complexes represents a functionally distinct PLI binding pocket (as codified by ECOD domains \citep{cheng2014ecod}, see \cite{corso_2024_10656052} for more details) compared to data deposited in the PDB before 2019. As such, it is likely that Chai-1, Boltz-1, and AF3 are ``overfitted'' to the most common types of PLI structures in the PDB and may overlook several uncommon types of PLI binding pockets present in nature.

\subsection{PoseBusters Benchmark results}\label{section:posebusters_benchmark_results}

\begin{figure}
  \centering
  \includegraphics[width=\linewidth]{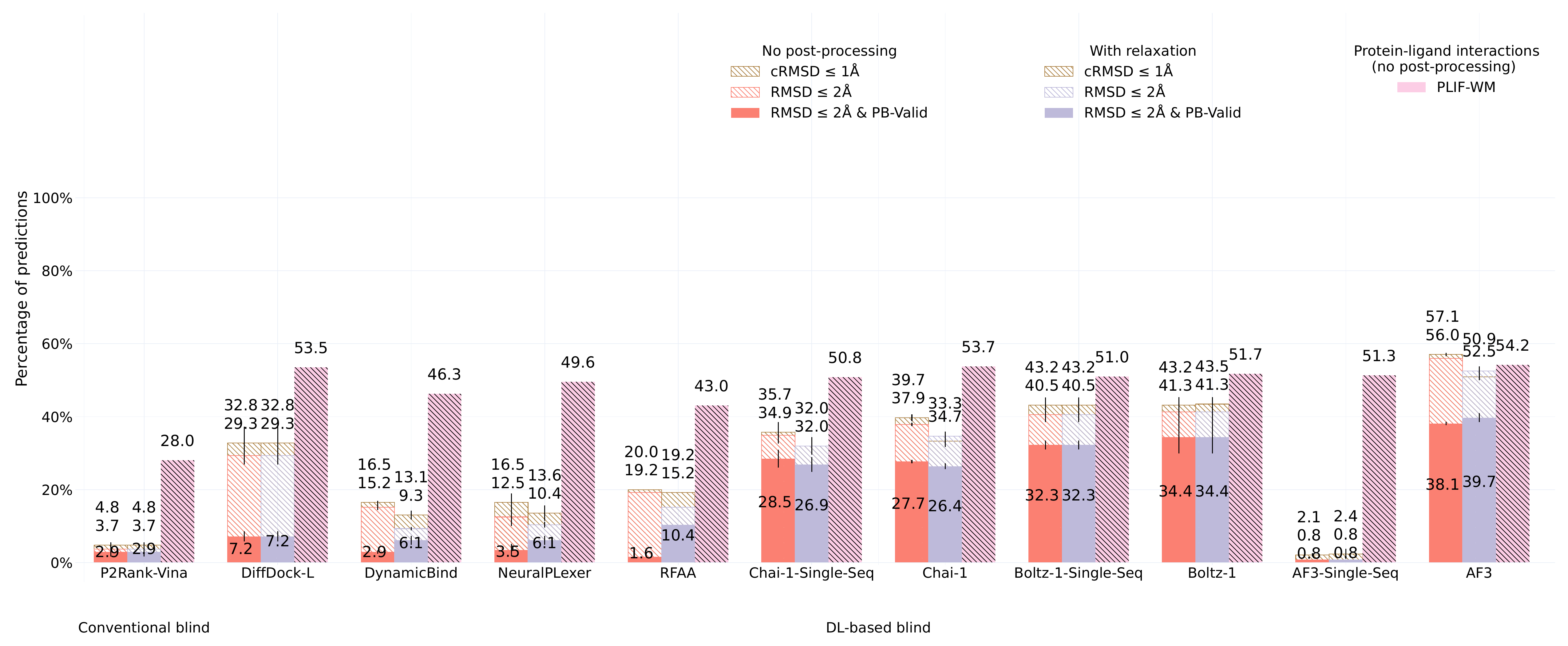}
  \caption{PoseBusters Benchmark primary ligand docking success rates (n=130/308 protein-ligand complexes). Data are presented as mean values +/- standard deviations over three independent predictions for each complex.}
  \label{figure:posebusters_benchmark_primary_ligand_relaxed_rmsd_lt2_bar_chart}
\end{figure}

With approximately half of its PLI structures deposited in the PDB after AF3 and Boltz-1's maximum-possible training data cutoff of September 30, 2021 (n=308 total protein-ligand complexes, filtered to n=130 for subsequent analyses), the PoseBusters Benchmark dataset's results, presented in Figure \ref{figure:posebusters_benchmark_primary_ligand_relaxed_rmsd_lt2_bar_chart}, indicate once again that DL co-folding methods achieve top performance compared to conventional and DL docking baseline methods. Nonetheless, we observe an interesting phenomenon whereby Boltz-1 strikes a better balance of structural and chemical accuracy and chemical specificity compared to Chai-1, even without input MSAs, potentially suggesting that, for this dataset, Boltz-1 achieves binding pose generalization approaching that of AF3. This notion is reinforced by the observation that, for this dataset, the single-sequence version of AF3 (again) displays significant degradations in its overall performance. These observations highlight the importance in future work of carefully studying why and how the \textit{training} of biomolecular structure generative models can be influenced to varying degrees by the availability and composition of diverse input MSAs.

\subsection{CASP15 results}\label{section:casp15_results}

\begin{figure}
  \centering
  \includegraphics[width=\linewidth]{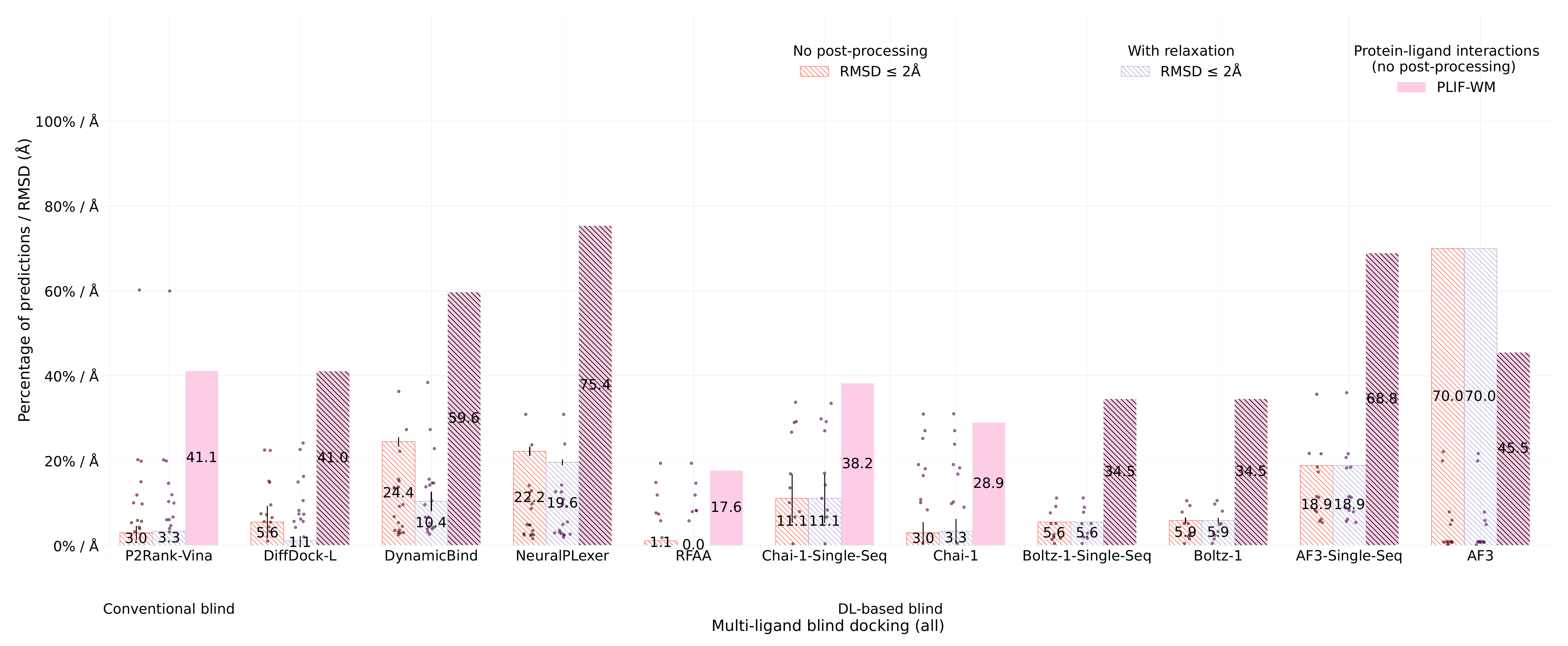}
  \caption{CASP15 multi-ligand docking success rates (n=13 protein-ligand complexes). Data are presented as mean values +/- standard deviations over three independent predictions for each complex.}
  \label{figure:casp15_all_multi_ligand_relaxed_rmsd_lt2_bar_chart}
\end{figure}

\begin{extfigure}
  \centering
  \includegraphics[width=\linewidth]{images/Morehead_ED_Fig1.jpg}
  \caption{CASP15 single-ligand docking success rates (n=6 protein-ligand complexes). Data are presented as mean values +/- standard deviations over three independent predictions for each complex.}
  \label{extfigure:casp15_all_single_ligand_relaxed_rmsd_lt2_bar_chart}
\end{extfigure}

As a new dataset of novel and challenging PLI complexes on which no method has been trained, the CASP15 dataset's multi-ligand results (n=13 protein-ligand complexes), illustrated in Figure \ref{figure:casp15_all_multi_ligand_relaxed_rmsd_lt2_bar_chart}, indicate that most methods fail to adequately generalize to multi-ligand prediction targets, yet AF3 stands out in this regard (only) when provided input MSAs. As many of these CASP15 multi-ligand targets represent large, highly symmetric protein complexes, it is likely that additional evolutionary information in the form of MSAs has improved AF3's ability to predict higher-order protein-protein interactions for these targets, yet interestingly its improved rate of structural accuracy comes at the cost of its protein-\textit{ligand} chemical specificity (in comparison to its single-sequence results). For the CASP15 dataset's single-ligand (i.e., primary ligand) results (n=6 protein-ligand complexes) presented in Extended Data Figure \ref{extfigure:casp15_all_single_ligand_relaxed_rmsd_lt2_bar_chart}, this trend is subverted in that conventional docking and open-source DL co-folding methods such as AutoDock Vina, NeuralPLexer, and Boltz-1 outperform all other recent DL co-folding methods in modeling crystalized PLIFs while achieving comparable rates of structural accuracy. Given the small size of the CASP15 dataset, it is reasonable to conclude that DL methods, in particular some of the latest co-folding methods, \textit{may} be challenged to predict protein-ligand complexes containing novel PLIs. In the following Section \ref{section:exploratory_analyses_of_results}, we will explore this latter point in greater detail by analyzing the protein-ligand binding similarities between common PDB training data and this benchmark's evaluation datasets.

\begin{extfigure}
  \centering
  \includegraphics[width=\linewidth]{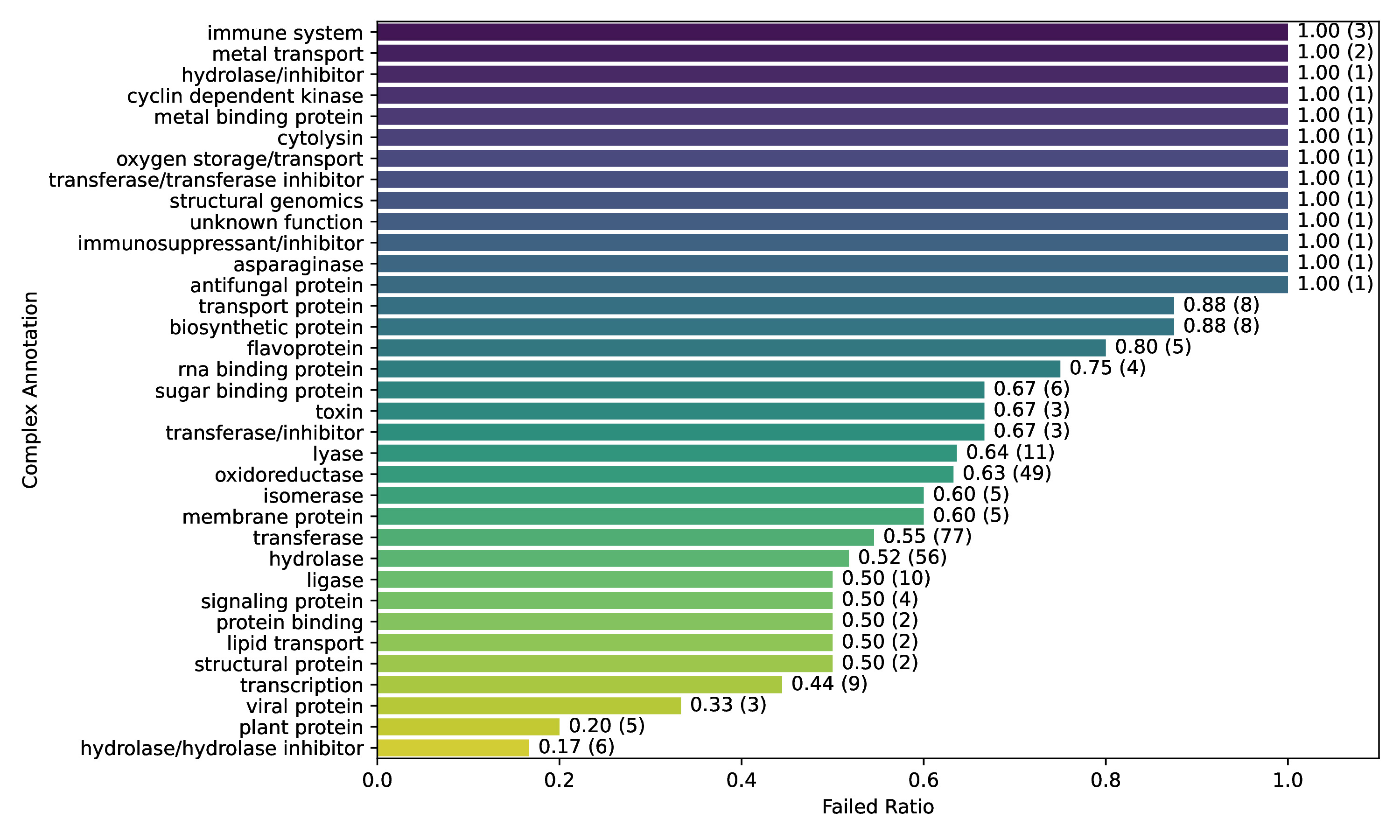}
  \caption{Function annotations of the PLI complexes all methods mispredicted (n=177 protein-ligand complexes).}
  \label{extfigure:failed_complexes_functional_keywords_1}
\end{extfigure}

\subsection{Exploratory analyses of results}\label{section:exploratory_analyses_of_results}
In this section, we explore a range of questions to study the common ``failure'' modes of the baseline methods included in this work, to outline new directions for future research and development efforts in drug discovery. \\

\begin{extfigure}
  \centering
  \includegraphics[width=\linewidth]{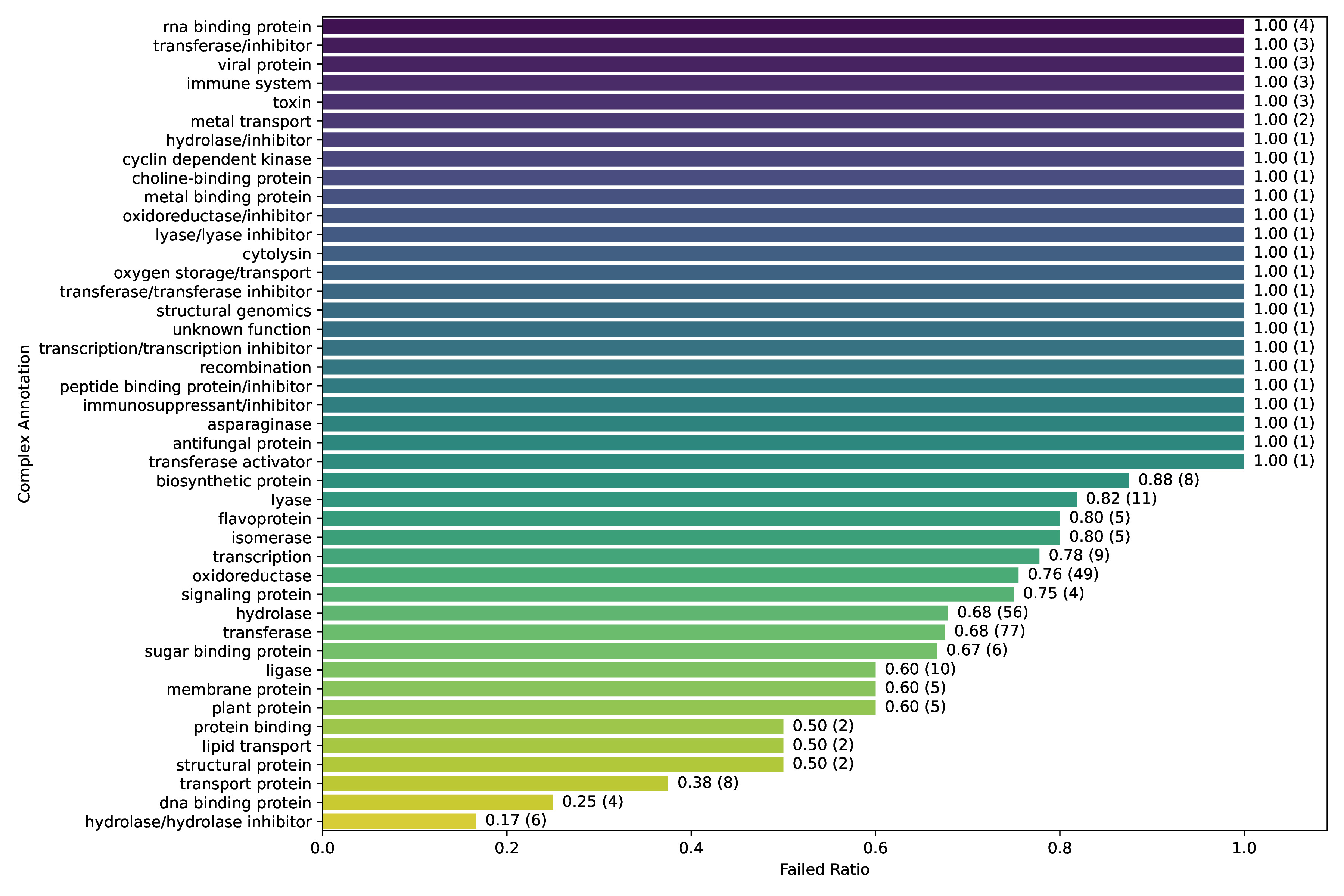}
  \caption{Function annotations of the PLI complexes AF3 mispredicted (n=221 protein-ligand complexes).}
  \label{extfigure:failed_af3_complexes_functional_keywords}
\end{extfigure}

\textbf{Research Question 1}: What are the most common types of protein-ligand complexes that \textit{all} baseline methods fail to predict?

$\rightarrow$ To address this query, we first collect all ligand pose predictions that no method could predict with structural and chemical accuracy (according to the metric RMSD $\leq$ 2 \AA\ \& PB-Valid). For each of these ``failed'' ligand poses, we retrieve the PDB's functional annotation of the protein in complex with this ligand and construct a histogram to visualize the frequency of these (failed complex) annotations. The results of this analysis are presented in Extended Data Figure \ref{extfigure:failed_complexes_functional_keywords_1}, in which we see that immune system proteins, metal transport proteins, biosynthetic proteins, flavoproteins, RNA binding proteins, and oxidoreductases are commonly mispredicted by all baseline methods such as Chai-1 and RoseTTAFold-All-Atom (RFAA) \citep{krishna2024generalized}, suggesting these classes of proteins may be largely unaddressed by the most recent DL methods for PLI structure prediction. To illuminate potential future research directions, in the next analysis, we investigate whether this pattern persists specifically for AF3, the most accurate DL co-folding method \textit{according to our benchmarking results}. \\

\textbf{Research Question 2}: What are the most common types of protein-ligand complexes that highly-accurate DL \textit{co-folding} methods such as AF3 fail to predict?

$\rightarrow$ For this follow-up question, we link all of AF3's failed ligand predictions with corresponding protein function annotations available in the PDB to understand which types of PLI complexes AF3 finds the most difficult to predict (to understand its predictive coverage of important molecular functions). Similar to the answer to our first research question, Extended Data Figure \ref{extfigure:failed_af3_complexes_functional_keywords} shows that, in order of difficulty, AF3 is most challenged to produce ligand poses of high structural and chemical accuracy for ligand-bound RNA binding proteins, immune system proteins, metal transport proteins, biosynthetic proteins, lyases, flavoproteins, and oxidoreductases. As several of these classes of proteins have not been well represented in the PDB over the last 50 years (e.g., immune system and biosynthetic proteins), in future work, it will be important to ensure that either the performance of new DL methods for PLI structure prediction is expanded to support accurate modeling of these uncommon types of ligand-bound proteins or a broadly applicable fine-tuning method for uncommon types of interactions is proposed. \\

\textbf{Research Question 3}: Is \textit{lack} of protein-ligand binding pose homology to PDB training data (inversely) correlated with the prediction accuracy of each method?

\begin{extfigure}
  \centering
  \includegraphics[width=\linewidth]{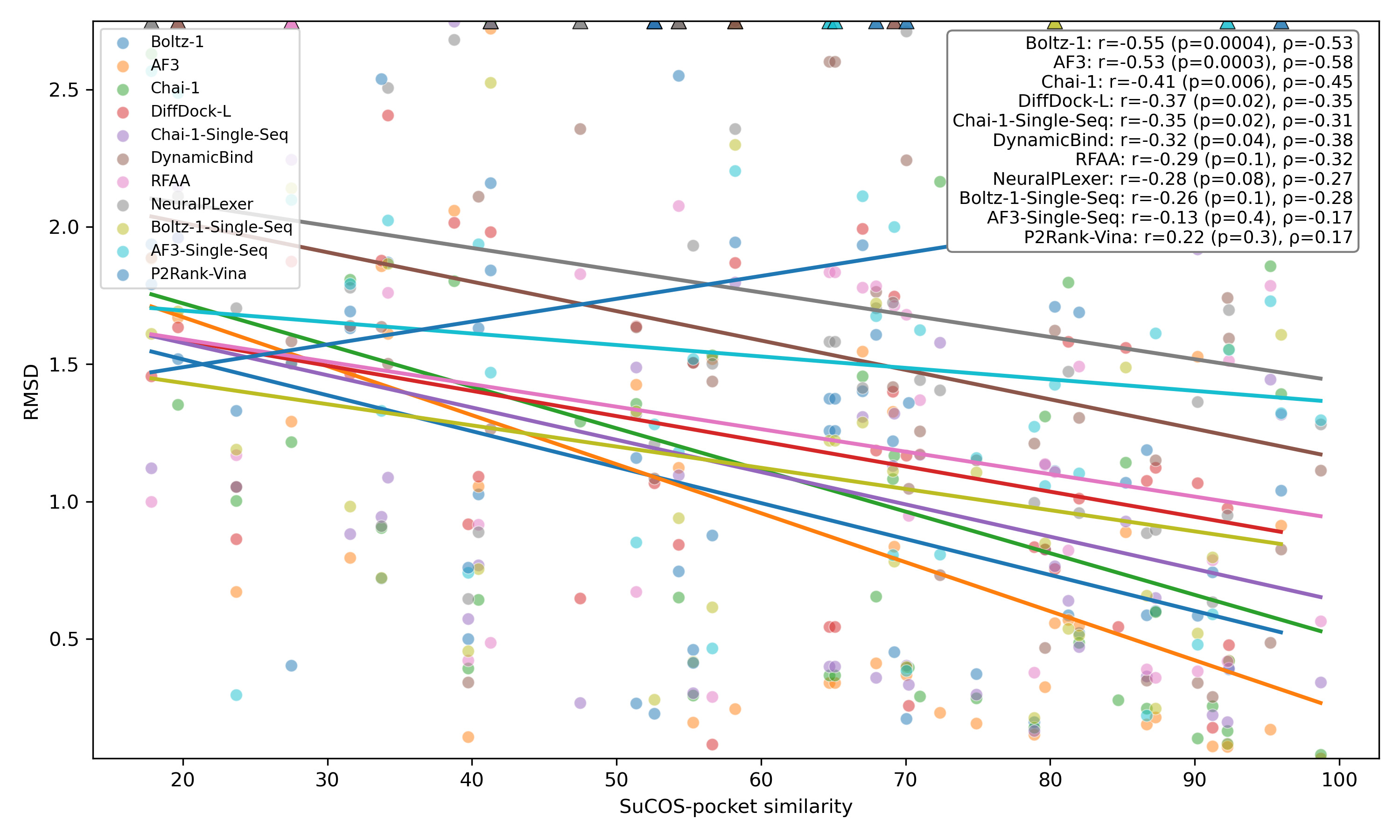}
  \caption{Linear regression lines between each method's PoseBusters Benchmark prediction accuracy (RMSD) \textit{after optimal alignment of the predicted ligand pose to the ground-truth ligand pose} and each complex's training set similarity in terms of SuCOS-pocket similarity (n=41 protein-ligand complexes for each method). Pearson (r, with p-values) and Spearman ($\rho$) correlation coefficients were calculated to quantify the relationship between each method's prediction RMSD and its target's training set similarity using two-sided tests from SciPy. Reported p-values are unadjusted, as no correction for multiple comparisons was applied.}
  \label{extfigure:posebusters_benchmark_methods_generalization_analysis}
\end{extfigure}

\begin{extfigure}
  \centering
  \includegraphics[width=\linewidth]{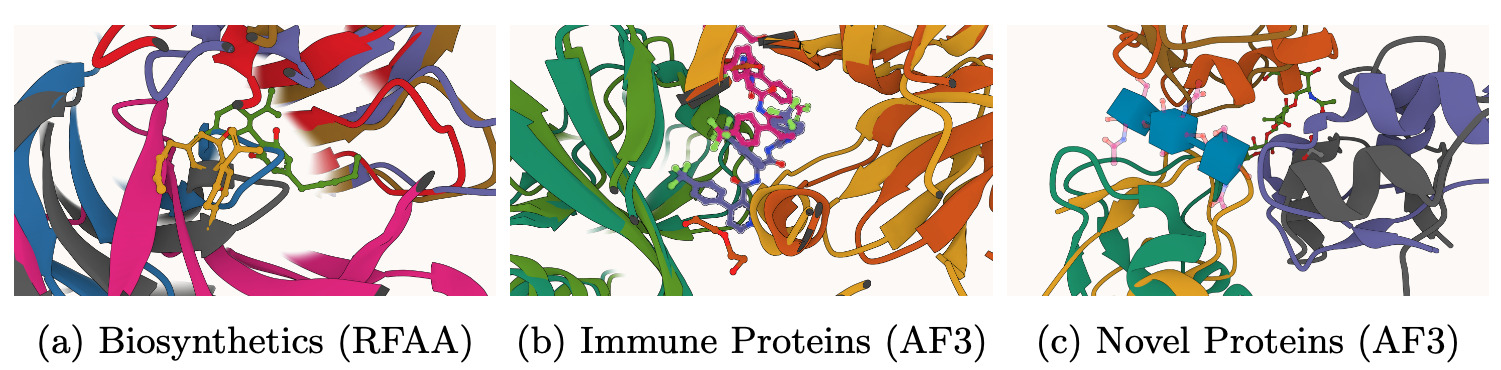}
  \caption{Examples of baseline methods' three failure modes (a: biosynthetics, b: immune proteins, c: novel proteins) discussed in this work.}
  \label{extfigure:rq_failure_modes_analysis}
\end{extfigure}

$\rightarrow$ To understand the impact of protein-ligand binding pose similarity on the performance of each baseline method, we use the PLINDER data resource \citep{durairaj2024plinder} to identify (n=41/130 protein-ligand complexes) cluster representatives of the PoseBusters Benchmark dataset based on the product of each complex's ligand (structure and feature-based) Combined Overlap Score (SuCOS) \citep{leung2019sucos} and its protein binding pocket's (structure and sequence-based) similarity \citep{vskrinjar2025have}, as none of this subset's prediction targets are contained in any method's training dataset. For these cluster representatives, with SciPy 1.15.1 \citep{virtanen2020scipy} we then calculate the Pearson and Spearman correlations (and p-values) between each method's complex prediction accuracy (i.e., ligand pose RMSD) \textit{after optimal alignment of the predicted ligand pose to the ground-truth ligand pose} and the complex's maximum SuCOS-binding pocket-based similarity to any complex deposited in the PDB before AF3's training dataset cutoff of September 30, 2021. Extended Data Figure \ref{extfigure:posebusters_benchmark_methods_generalization_analysis} reveals that \textit{all} DL methods' performance (according to this definition) is correlated with a complex's similarity to common PDB training data, with (MSA-based) Boltz-1, AF3, and Chai-1 exhibiting the strongest and most statistically significant (p $< 0.05$) correlations. Like concurrent work assessing the performance of co-folding methods in novel prediction settings \citep{vskrinjar2025have}, our findings suggest that although the current generation of deep learning models (\textit{both} docking and co-folding methods) for protein-ligand docking and structure prediction can occasionally make accurate ligand conformation predictions for truly novel (SuCOS-pocket similarity $<$ 30) protein-ligand complexes, such methods rely (at least in large part) on \textit{recapitulating} protein-ligand binding patterns seen during training to make accurate predictions for unseen complexes (n.b., while interestingly for conventional docking methods such as P2Rank-Vina, this trend is not observed). We conclude our quantitative analyses with an illustration of the different failure modes of each baseline method, as depicted in Extended Data Figure \ref{extfigure:rq_failure_modes_analysis}. In this figure, we illustrate that DL methods such as RFAA and AF3 commonly struggle to accurately predict the structure of ligand-binding biosynthetic and immune system proteins, suggesting that these (uncommon) types of PLIs are not well addressed by the current generation of DL-based structure prediction methods, suggesting future research opportunities for interaction-specific modeling (e.g., through the use of fine-tuning or preference optimization). \\

\begin{table}[!h]
  \caption{\textsc{PoseBench} evaluation datasets of protein-(multi-)ligand structures.}
  \label{table:preprocessed_datasets}
  \centering
  \begin{tabularx}{\textwidth}{lXXXX}
    \toprule
    & Astex Diverse & DockGen-E & PoseBusters Benchmark & CASP15 \\
    \midrule
    Ligand Type & Primary & Primary & Primary & Multi \\
    Source & \citep{hartshorn2007diverse} &  & \citep{buttenschoen2024posebusters} &  \\
    Size (Total \# Ligands) & 85 & 122 & 130/308 & 102 (across 19 complexes) $\rightarrow$ 6 (13) single (multi)-ligand complexes \\
    Training Data Homology & High & Moderate & Low & Low \\
    Top-Ranked Method (w/ MSAs)  &   AF3   & AF3  &  AF3   & AF3 \\
    Top-Ranked Method (w/o MSAs)  &   Boltz-1   & Chai-1  &  Boltz-1   & NeuralPLexer \\
    \bottomrule
  \end{tabularx}
\end{table}

\section{Conclusions}\label{section:conclusions}
In this work, we have introduced \textsc{PoseBench}, a unified, broadly applicable benchmark and toolkit for studying the performance of methods for protein-ligand docking and structure prediction. Benchmarking results with \textsc{PoseBench}, summarized in Table \ref{table:preprocessed_datasets}, suggest that DL co-folding methods generally outperform conventional and DL docking baselines yet remain challenged to predict complexes containing novel protein-ligand binding poses, with AF3 performing best overall when deep MSAs are available for a target protein, regardless of the availability of homologous proteins. Further, we find that several DL methods face difficulties balancing the structural accuracy of their predicted poses with the chemical specificity of their induced protein-ligand interactions, highlighting that future methods may benefit from the introduction of \textit{physico-chemical} loss functions or sampling techniques to bridge this performance gap. Lastly, we observe that some (but not all) DL co-folding methods are highly dependent on the availability of diverse input MSAs to achieve high structural prediction accuracy (e.g., AF3 but not Chai-1 or Boltz-1), underscoring the need in future work to elucidate the impact of the availability of MSAs and protein language model embeddings \citep{lin2023evolutionary} on the training dynamics of biomolecular structure prediction methods. As a publicly available resource, \textsc{PoseBench} is flexible to accommodate new datasets, methods, and analyses for protein-ligand docking and structure prediction.

\section{Methods}\label{section:methods}

\subsection{\textsc{PoseBench}}\label{section:posebench}

The overall goal of \textsc{PoseBench}, our newly designed benchmark for protein-ligand docking and structure prediction, is to provide the research community with a centralized resource with which one can systematically measure, in a variety of macromolecular contexts, the methodological advancements of new conventional and DL methods proposed for this domain. In the following sections, we describe \textsc{PoseBench}'s design and composition (as portrayed in Figure \ref{figure:posebench}) and how we have used \textsc{PoseBench} to evaluate several recent DL docking and co-folding methods (as well as a strong conventional baseline algorithm) for protein-ligand structure modeling.

\subsection{Benchmark datasets}\label{section:benchmark_datasets}

As shown in Table \ref{table:preprocessed_datasets}, \textsc{PoseBench} provides users with \textit{broadly applicable}, preprocessed versions of four datasets with which to evaluate existing or new protein-ligand structure prediction methods: Astex Diverse \citep{hartshorn2007diverse}, PoseBusters Benchmark \citep{buttenschoen2024posebusters}, and the new DockGen-E and CASP15 PLI datasets that we have manually curated in this work.

\textbf{Astex Diverse dataset.} The Astex Diverse dataset is a collection of 85 PLI complexes composed of various drug-like molecules and cofactors known to be of pharmaceutical or agrochemical interest, where a primary (representative) ligand is annotated for each complex. This dataset can be considered an easy benchmarking dataset for methods trained on recent data contained in the PDB in that most of its complexes (deposited in the PDB up to 2007) are known to overlap with the commonly used PDBBind 2020 (time-split) training dataset \citep{liu2017forging, stark2022equibind} containing complexes deposited in the PDB before 2019. As such, including this dataset for benchmarking allows one to estimate the \textit{breadth} of a method's structure prediction capabilities for important primary ligand protein complexes represented in the PDB.

To perform unbound (apo) protein-ligand docking with this dataset, we used AF3 to predict the structure of each of its protein complexes, with all ligands and cofactors excluded. We then optimally aligned these predicted protein structures to the corresponding crystal (holo) PLI complex structures using a PLI binding site-focused structural alignment performed using PyMOL \citep{delano2002pymol}, where each binding site is defined as all amino acid residues containing crystallized heavy atoms that are within 10 \AA\ of any crystallized ligand heavy atom. To enable the broad availability of \textsc{PoseBench}'s benchmark datasets in both commercial and academic settings, we also provide unbound (apo) protein structures predicted using the MIT-licensed ESMFold model \citep{lin2023evolutionary}, although in Section \ref{section:results_and_discussion} we report results using AF3's predicted structures as the default data source. We further note that on average across all benchmark datasets and methods, AF3's predicted structures improve baseline docking methods' structural accuracy rates by 5-10\%.

\textbf{PoseBusters Benchmark dataset.} Version 2 of the popular PoseBusters Benchmark dataset \citep{buttenschoen2024posebusters}, which we adopt in this work, contains 308 recent primary ligand protein complexes deposited in the PDB from 2019 onwards. Accordingly, in contrast to Astex Diverse, this dataset can be considered a moderately difficult benchmark dataset for baseline methods, since many of its complexes do not directly overlap with the most commonly used PDB-based training data. Important to note is that, among all baseline methods, AF3 and Boltz-1 used the most recent PDB training data cutoff of September 30, 2021, which motivated us to report the results in Section \ref{section:posebusters_benchmark_results} for only the subset of PoseBusters Benchmark complexes (n=130 protein-ligand complexes) deposited in the PDB after this date. Like Astex Diverse, for the PoseBusters Benchmark dataset, we used AF3 (and ESMFold) to predict the \textit{apo} protein structures of each of its complexes and then performed our PyMOL-based structural binding site alignments.

\textbf{DockGen-E dataset.} The original DockGen dataset \citep{corso2024deep} contains 189 diverse primary ligand protein complexes, each representing a functionally distinct type of PLI binding pocket according to ECOD domain partitioning \citep{cheng2014ecod, corso_2024_10656052}. Consequently, this dataset can be considered \textsc{PoseBench}'s most difficult primary ligand dataset to model since its PLI binding sites are distinctly uncommon compared to those frequently found in the training datasets of all baseline methods, though it is important to note that these original DockGen complexes were deposited in the PDB from 2019 onward, making this benchmarking dataset partially overlap with the training datasets of baseline DL co-folding methods such as Chai-1, Boltz-1, and AF3. Nonetheless, in line with our initial hypotheses, the benchmarking results in Section \ref{section:results_and_discussion} demonstrate that no baseline method can adequately predict the PLI binding sites and ligand poses represented by this bespoke subset of the PDB, suggesting that \textit{all} baseline DL methods have yet to learn \textit{broadly applicable} representations of protein-ligand binding.

Unfortunately, the original DockGen dataset contains only the primary protein chains representing each novel binding pocket after filtering out all non-interacting chains and cofactors in a given biological assembly (bioassembly), which considerably \textit{reduces} the biophysical context provided to baseline methods to make reasonable predictions. As such, we argue for the need to construct a new dataset that challenges baseline methods (in particular DL co-folding methods) to predict full bioassemblies containing novel PLI binding pockets, which we address with our enhanced version of DockGen called DockGen-E.

To construct DockGen-E, we collected the original DockGen dataset's PLI binding pocket annotations for each complex. We then retrieved the corresponding first bioassembly listed in the PDB to obtain each PDB entry's biologically \textit{relevant} complex, filtering out DockGen complexes for which the first bioassembly could not be mapped to its original PLI binding pocket annotation (which indicates these original DockGen PLI binding pockets were initially not derived from the PDB's corresponding first bioassembly). This procedure left 122 biologically relevant assemblies remaining for benchmarking. Like Astex Diverse and PoseBusters Benchmark, for DockGen-E, we then used AF3 (and ESMFold) to predict the unbound (apo) protein structures of each complex in the dataset and structurally aligned the predicted protein structures to their corresponding crystallized PLI binding sites using PyMOL.

\textbf{CASP15 dataset.} To assess the \textit{multi}-primary ligand (i.e., multi-ligand) modeling capabilities of recent methods for protein-ligand docking and structure prediction, with \textsc{PoseBench}, we introduce a preprocessed, DL-ready version of the CASP15 PLI dataset debuted as a first-of-its-kind prediction category in the 15th Critical Assessment of Techniques for Structure Prediction (CASP) competition held in 2022 \citep{robin2023assessment}. The CASP15 PLI dataset is originally comprised of 23 protein-ligand complexes released in the PDB from 2022 onward, where we subsequently filter out 4 complexes based on (1) whether the CASP organizers ultimately assessed predictions for the complex and (2) whether they are nucleic acid-ligand complexes with no interacting protein chains. The 19 remaining PLI complexes, which contain a total of \underline{102} (fragment) ligands, consist of a variety of ligand types including single-atom (metal) ions and large drug-sized molecules with up to 92 atoms in each (fragment) ligand. As such, this dataset is appropriate for assessing how well structure prediction methods can model interactions between different (fragment) ligands in the same complex, which can yield insights into the inter-ligand steric clash rates of each method. As with all other benchmark datasets, we used AF3 (and ESMFold) to predict the unbound (apo) structure of each protein complex in the dataset and then performed a PyMOL-based structural alignment of the corresponding PLI binding sites.

\textbf{PLI similarity analysis between datasets.} For an investigation of the similarity of PLIs represented in each dataset, in Supplementary Appendix \ref{section:appendix_analysis_of_protein_ligand_interactions}, we analyze the different types and frequencies of common, ProLIF-annotated protein-ligand binding pocket interactions \citep{bouysset2021prolif} natively found within the common PDBBind 2020 training dataset and the Astex Diverse, PoseBusters Benchmark, DockGen-E, and CASP15 datasets, respectively, to quantify the diversity of the (predicted) interactions each dataset can be used to evaluate. In short, we find that the DockGen-E and CASP15 benchmark datasets are the \textit{most dissimilar} compared to the common PDBBind 2020 training dataset, further illustrating the unique PLI modeling challenges offered by these evaluation datasets.

\subsection{Formulated tasks}\label{section:prediction_tasks}
In this work, we developed \textsc{PoseBench} to focus our analysis on the behavior of different conventional and DL methods for protein-ligand structure prediction in a variety of macromolecular contexts (e.g., with or without inorganic cofactors present). With this goal in mind, below we formalize the structure prediction tasks currently available with \textsc{PoseBench}, with its source code flexibly designed to accommodate new tasks in future work.

\textbf{Primary ligand blind docking.} For primary ligand blind docking, each baseline method is provided with a complex's (multi-chain) protein sequence and an optional predicted (apo) protein structure as input along with its corresponding (fragment) ligand SMILES strings, where fragment ligands include the \textit{primary} binding ligand to be scored as well as all cofactors present in the corresponding crystal structure. In particular, no knowledge of the complex's PLI binding pocket is provided to evaluate how well each method can (1) identify the correct PLI binding pockets and (2) correct ligand poses within each pocket (3) with high chemical validity and (4) specificity for the pockets' amino acid residues. After all fragment ligands are predicted, \textsc{PoseBench} extracts each method's prediction of the primary binding ligand and reports evaluation results for these primary predictions.

\textbf{Multi-ligand blind docking.} For multi-ligand blind docking, each baseline method is provided with a complex's (multi-chain) protein sequence and an optional predicted (apo) protein structure as input along with its corresponding (fragment) ligand SMILES strings. As in primary ligand blind docking, no knowledge of the PLI binding pockets is provided, which offers the opportunity to evaluate not only PLI binding pocket and conformation prediction accuracy but, in the context of multi-binding ligands, also inter-ligand steric clash rates.

\subsection{Metrics}\label{section:metrics}
\textbf{Traditional metrics.} For \textsc{PoseBench}, we reference two key metrics in the field of structural bioinformatics: the root-mean-square deviation (RMSD) and local Distance Difference Test (lDDT) \citep{mariani2013lddt}. The RMSD between a predicted 3D conformation (with atomic positions $\hat{x}_{i}$ for each of the molecule's $n$ heavy atoms) and the ground-truth (crystal structure) conformation ($x_{i}$) is defined as:

\begin{equation}
\label{equation:rmsd_definition}
    \text{RMSD} = \sqrt{\frac{1}{n} \sum_{i = 1}^{n} \lVert \hat{x}_{i} - x_{i} \rVert^{2}}.
\end{equation}

The lDDT score, which is commonly used to compare predicted and ground-truth protein 3D structures, is defined as:

\begin{equation}
\label{equation:lddt_definition}
    \text{lDDT} = \frac{1}{N} \sum_{i=1}^{N} \frac{1}{4} \sum_{k=1}^{4} \left( \frac{1}{|\mathcal{N}_i|} \sum_{j \in \mathcal{N}_i} \Theta(|\hat{d}_{ij} - d_{ij}| < \Delta_k) \right),
\end{equation}
where $N$ is the total number of heavy atoms in the ground-truth structure; $\mathcal{N}_{i}$ is the set of neighboring atoms of atom $i$ within the inclusion radius $R_{o} = 15\ \AA$ in the ground-truth structure, excluding atoms from the same residue; $\hat{d}_{ij}$ ($d_{ij}$) is the distance between atoms $i$ and $j$ in the predicted (ground-truth) structure; $\Delta_{k}$ are the distance tolerance thresholds (i.e., 0.5 \AA, 1 \AA, 2 \AA, and 4 \AA); $\Theta(x)$  is a step function that equals 1 if x is true, and 0 otherwise; and $|\mathcal{N}_{i}|$ is the number of neighboring atoms for atom $i$. As originally proposed by \citet{robin2023assessment}, in this study, we adopt the PLI-specific variant of lDDT for scoring \textit{multi}-ligand complexes, which calculates lDDT scores to compare predicted and ground-truth protein-(multi-)ligand complex structures following optimal (chain-wise and residue-wise) structural alignment of the predicted and ground-truth PLI binding pockets.

Lastly, we also measure the molecule validity rates of each predicted PLI complex pose using the PoseBusters software suite (i.e., PB-Valid) \citep{buttenschoen2024posebusters}. This suite runs several important chemical and structural sanity checks for each predicted pose including energy ratio inspection and geometric (e.g., flat aliphatic ring) assertions which provide a secondary filter of accurate poses that are also chemically and structurally meaningful.

\textbf{New metrics.} The RMSD, lDDT, and PB-Valid metrics of a protein-ligand binding structure provide useful characterizations of how accurate and reasonable a predicted pose is. However, a key limitation of these metrics is that they do not measure how well a predicted pose resembles a native pose when comparing their induced PLIFs. Recently, \cite{errington2024assessing} introduced a complementary benchmarking metric, PLIF-valid, assessing DL methods' recovery rates of known PLIs. However, this metric only reports a strict recall rate of each method's interaction types rather than a continuous measure of how well each method's interactions match the \textit{distribution} of crystalized PLIs. Moreover, in drug discovery, a primary concern when designing new drug candidates is ensuring they produce \textit{amino acid-specific} types of interactions (and not others), hence we desire each baseline method to recall the correct types of PLIs for each pose and to avoid predicting (i.e., hallucinating) types of interactions that are not natively present. Consequently, we argue that an ideal PLI-aware benchmarking metric is a single continuous metric that assesses the recall and precision of a method's predicted \textit{distribution} of \textit{amino acid-specific} PLIFs. To this end, we propose two new benchmarking metrics, PLIF-EMD and PLIF-WM.

For each PLI complex, PLIF-EMD measures the Earth mover's distance (EMD) \citep{rubner2000earth} between a method's predicted histogram of PLI type counts $u$ (specific to each type of interaction) and the corresponding native histogram $v$, where each histogram of interaction type counts is represented as a 1D discrete distribution. Formally, this equates to computing the Wasserstein distance between these two 1D distributions $u$ and $v$ as
\begin{equation}
    \label{equation:plif_emd_definition}
    \text{PLIF-EMD}\ :=\ l_{1}(u, v) = \inf_{\pi \in \Pi(u, v)} \int_{\mathbb{R} \times \mathbb{R}} |x - y| d\pi(x, y),
\end{equation}
where $\Pi(u, v)$ denotes the set of distributions on $\mathbb{R} \times \mathbb{R}$ whose marginals, $u$ and $v$, are on the first and second factors, respectively. To penalize a baseline method for producing non-native interaction types, we unify the bins in each histogram before converting them into 1D discrete representations. Namely, to perform this calculation, each PLI is first represented as a fingerprint tuple of $<$ligand type, amino acid type, interaction type$>$ as determined by the software tool ProLIF \citep{bouysset2021prolif} and then grouped to count each type of tuple to form a histogram. As such, a lower PLIF-EMD value implies a better continuous agreement between predicted and native interaction histograms. PLIF-WM, derived from PLIF-EMD, assesses the Wasserstein matching (WM) score of a pair of PLIF histograms. Specifically, to obtain a more benchmarking-friendly score ranging from 0 to 1 (higher is better), we define PLIF-WM as
\begin{equation}
    \label{equation:plif_wm_definition}
    \text{PLIF-WM}\ :=\ 1 - \frac{\text{PLIF-EMD} - \text{PLIF-EMD}_{\text{min}}}{\text{PLIF-EMD}_{\text{max}} - \text{PLIF-EMD}_{\text{min}}},
\end{equation}
where $\text{PLIF-EMD}_{\text{min}}$ and $\text{PLIF-EMD}_{\text{max}}$ denote the minimum (best) and maximum (worst) values of PLIF-EMD, respectively. As a metric normalized relative to each collection of the latest baseline methods, PLIF-WM allows one to quickly identify which of the latest methods has the greatest capacity to produce realistic distributions of PLIs. As a practical note, we use SciPy 1.15.1 \citep{virtanen2020scipy} to provide users of \textsc{PoseBench} with an optimized implementation of PLIF-EMD and thereby PLIF-WM.

\subsection{Baseline methods and experimental setup}
\label{section:baseline_methods_and_experimental_setup}

\textbf{Overview.} We designed \textsc{PoseBench} to answer specific modeling questions for PLI complexes such as (1) which types of methods (if any) can predict both common and uncommon PLI complexes with high structural and chemical accuracy and (2) which most accurately predict multi-ligand structures without steric clashes? In the following sections, we discuss which types of methods we evaluate in our benchmark and how we evaluate each method's predictions for PLI complex targets.

\textbf{Method categories.} As illustrated in Figure \ref{figure:posebench}, to explore a range of the most well-known or recent methods to date, we divide \textsc{PoseBench}'s baseline methods into one of three categories: (1) \underline{conventional} algorithms, (2) \underline{DL docking} algorithms, and (3) \underline{DL co-folding} algorithms.

As a representative algorithm for \underline{conventional} protein-ligand docking, we pair AutoDock Vina (v1.2.5) \citep{trott2010autodock} for molecular docking with P2Rank for protein-ligand binding site prediction \citep{krivak2018p2rank} to form a strong conventional (blind) docking baseline (P2Rank-Vina) for comparison with DL methods. To represent \underline{DL docking} methods, we include DiffDock-L \citep{corso2024deep} for docking with static protein structures and DynamicBind \citep{lu2024dynamicbind} for flexible docking. Lastly, to represent some of the latest \underline{DL co-folding} methods, we include NeuralPLexer \citep{qiao2024state}, RFAA \citep{krishna2024generalized}, Chai-1 \citep{chai2024chai}, Boltz-1 \citep{wohlwend2024boltz} (versus Boltz-2 \citep{passaro2025boltz} for sake of time-split benchmarking validity), and AF3 \citep{abramson2024accurate}. For interested readers, each method's input and output data formats are described in Supplementary Appendix \ref{section:appendix_additional_method_descriptions}.

\textbf{Prediction and evaluation procedures.} The PLI complex structures each method predicts are subsequently evaluated using different structural and chemical accuracy and molecule validity metrics depending on whether the targets are primary or multi-ligand complexes. In Section \ref{section:metrics}, we provide formal definitions of \textsc{PoseBench}'s evaluation metrics. Note that if a method's prediction raises any errors in subsequent scoring stages (e.g., due to missing entities or formatting violations), the prediction is excluded from the evaluation.

\textbf{Primary ligand evaluation.} For primary ligand targets, we report each method's percentage of (top-1) ligand conformations within 2 \AA\ of the corresponding crystal ligand structure (RMSD $\leq$ 2 \AA), using 1 \AA\ to instead assess whether the predicted ligand's heavy atom centroid (i.e., binding pocket) was correct (cRMSD $\leq$ 1 \AA), as well as the percentage of such "correct" ligand conformations that are also considered to be chemically and structurally valid according to the PoseBusters software suite \citep{buttenschoen2024posebusters} (RMSD $\leq$ 2 \AA\ \& PB-Valid). Importantly, as described in Section \ref{section:metrics}, we also report each method's new PLIF-WM scores to study the relationship between its structural accuracy and chemical specificity.

\textbf{Multi-ligand evaluation.} Similar to the protein-ligand scoring procedure employed in the CASP15 competition \citep{robin2023assessment}, for multi-ligand targets, we report each method's (top-1) percentage of "correct" (binding site-superimposed) ligand conformations (RMSD $\leq$ 2 \AA) as well as violin plots of the RMSD and PLI-specific lDDT scores of its protein-ligand conformations across all (fragment) ligands within the benchmark's multi-ligand complexes (see Supplementary Appendix \ref{section:appendix_additional_results} for these plots). Notably, this latter metric, referred to as lDDT-PLI, allows one to evaluate specifically how well each method can model protein-ligand structural interfaces. Additionally, we report each method's PB-Valid rates (calculated once for each multi-ligand complex) and PLIF-WM scores.

\section{Data availability}
The \textsc{PoseBench} datasets and benchmark results are available at \href{https://zenodo.org/records/19138652}{https://zenodo.org/records/19138652} under a Creative Commons Attribution 4.0 International Public License \citep{morehead_2025_19138652}, with further licensing discussed in Supplementary Appendix \ref{section:appendix_availability} and detailed dataset documentation (e.g., of AF3's predicted protein structure accuracy) provided in Supplementary Appendix \ref{section:appendix_documentation_for_datasets}.

\section{Code availability}
The \textsc{PoseBench} codebase, documentation, and tutorial notebooks are available at \href{https://github.com/BioinfoMachineLearning/PoseBench}{https://github.com/BioinfoMachineLearning/PoseBench} under a permissive MIT license \citep{alex_morehead_2025_19140262}, with further licensing and broader impacts discussed in Supplementary Appendices \ref{section:appendix_availability} and \ref{section:appendix_broader_impacts}. In particular, the code makes use of the Python packages \texttt{hydra-core} 1.3.2, \texttt{biopandas} 0.5.1, \texttt{biopython} 1.79, \texttt{meeko} 0.6.0a3, \texttt{numpy} 1.26.4, \texttt{pandas} 2.0.0, \texttt{posebusters} 0.6.5, \texttt{posecheck} 1.1, \texttt{prolif} 2.1.0, \texttt{pypdb} 2.5, \texttt{rdkit} 2025.9.6, \texttt{scikit-learn} 1.7.2, \texttt{seaborn} 0.12.2, and \texttt{spyrmsd} 0.9.0.

\section{Acknowledgements}
This research used resources of the National Energy Research Scientific Computing Center (NERSC), a U.S. Department of Energy User Facility, using NERSC award DDR-ERCAP 0034574 awarded to AM. This work was also supported by a U.S. NSF grant (DBI2308699) and two U.S. NIH grants (R01GM093123 and R01GM146340) awarded to JC. Additionally, this work was performed using computing infrastructure provided by Research Support Services at the University of Missouri-Columbia (DOI: 10.32469/10355/97710). We would like to thank Martin Buttenschoen and Andriy Kryshtafovych for graciously allowing us in Figure \ref{figure:posebench} to parody the PoseBusters and CASP logos, respectively, using large language models hosted on CBorg, the multi-model AI portal of Lawrence Berkeley National Laboratory. We would like to specifically thank Martin Buttenschoen for their assistance in running the PoseBusters software suite for various PLI complexes. We would also like to thank Arian Jamasb for providing insightful feedback on an early version of this manuscript and Patrick Bryant for suggesting investigating the impact of sequence input ablations on model performance. Lastly, we would like to thank Zhuoran Qiao, Mia Rosenfeld, Feizhi Ding, Matthew Welborn, as well as anonymous reviewers for their helpful feedback during the development of the benchmark's alignment and scoring of PLI complex predictions.

\section{Author contributions statement}
AM and JC conceived the project. AM designed the experiments in this work, wrote their supporting code, analyzed their results, and described them in the current manuscript.  NG, JL, and PN helped run conventional and deep learning baseline methods for early versions of the experiments in this work and helped revise the current manuscript. JC provided funding, support, resources, important feedback, and manuscript revisions for this work.

\section{Competing interests statement}
The authors declare no competing interests.

\begin{appendices}

\appendixtoc
\clearpage

\section{Availability}\label{section:appendix_availability}

The \textsc{PoseBench} codebase and tutorial notebooks are available under an MIT license at \url{https://github.com/BioinfoMachineLearning/PoseBench}. Preprocessed datasets and benchmark method predictions and results are available on Zenodo
\citep{morehead_2025_19138652} under a CC-BY 4.0 license, of which the Astex Diverse and PoseBusters Benchmark datasets \citep{buttenschoen2024posebusters} and the DockGen-E dataset are associated with a CC-BY 4.0 license, and of which the CASP15 dataset \citep{robin2023assessment}, as a mixture of publicly and privately available resources, is partially licensed. In particular, 15 (4 single-ligand and 11 multi-ligand targets) of the 19 CASP15 protein-ligand interaction (PLI) complexes evaluated with \textsc{PoseBench} are publicly available, whereas the remaining 4 (2 single-ligand and 2 multi-ligand targets) are confidential and, for the purposes of future benchmarking and reproducibility, must be requested directly from the CASP organizers. Notably, the pre-holo-aligned protein structures predicted by AlphaFold 3 (AF3) for these four benchmark datasets (available on Zenodo \citep{morehead_2025_19138652}) must only be used in accordance with AF3's \href{https://github.com/google-deepmind/alphafold3/blob/main/WEIGHTS_TERMS_OF_USE.md}{Terms of Use}, whereas the pre-holo-aligned protein structures predicted by ESMFold for these four benchmark datasets (available on Zenodo \citep{morehead_2025_19138652}) are available under a permissive MIT license. Lastly, our use of the PoseBusters software suite for molecule validity checking is permitted under a BSD-3-Clause license.

\section{Broader impacts}\label{section:appendix_broader_impacts}

Our benchmark unifies protein-ligand structure prediction datasets, methods, and tasks to enable enhanced insights into the real-world utility of such methods for accelerated drug discovery and energy research. We acknowledge the risk that, in the hands of "bad actors", such technologies may be used with harmful ends in mind. However, it is our hope that efforts in elucidating the performance of recent protein-ligand structure prediction methods in various macromolecular contexts will disproportionately influence the positive societal outcomes of such research such as improved medicines and subsequent clinical outcomes as opposed to possible negative consequences such as the development of new bioweapons.

\section{Compute resources}\label{section:appendix_compute_resources}

To produce the results presented in this work, we ran a high performance computing sweep that concurrently utilized 12 80GB NVIDIA A100 GPUs for 14 days in total to run inference with each baseline method three times (where applicable), where each baseline deep learning (DL) method required approximately 24 hours of GPU compute to complete its inference runs (except for multiple sequence alignment (MSA)-dependent AF3 and RoseTTAFold-All-Atom (RFAA), which respectively took approximately 4 weeks and 2 weeks to finish their inference runs for each benchmark dataset). Notably, due to RFAA and AF3's significant storage requirements for running inference with their MSA databases, we utilized approximately 6 TB of solid-state storage space in total to benchmark all baseline methods. Lastly, in terms of CPU requirements, our experiments utilized approximately 64 concurrent CPU threads for AutoDock Vina inference (as an upper bound) and 60 GB of CPU RAM. Note that an additional 4-5 weeks of compute were spent performing initial (non-sweep) versions of each experiment during \textsc{PoseBench}'s initial phase of development.

\begin{table}
  \caption{The average runtime (in seconds) and peak memory usage (in GB) of each baseline method on a 25\% subset of the Astex Diverse dataset (using an NVIDIA 80GB A100 GPU for benchmarking). The symbol \textsc{-} denotes a result that could not be estimated. Where applicable, an integer enclosed in parentheses indicates the number of samples drawn from a particular baseline method.}
  \label{table:baseline_method_average_compute_resources}
  \centering
  \begin{tabular}{lccc}
    \toprule
    Method    & Runtime (s)     &   CPU Memory Usage (GB)   &   GPU Memory Usage (GB)    \\
    \midrule
    P2Rank-Vina (40)  &  1,283.70   &   9.62    &   0.00    \\
    DiffDock-L (5)  &   88.33  &   8.99    &   70.42   \\
    DynamicBind (5) & 146.99 &   5.26    &   18.91   \\
    NeuralPLexer (5)  &    29.10  &   11.19   &   31.00   \\
    RoseTTAFold-All-Atom (1)  & 3,443.63  &   55.75   &   72.79   \\
    Chai-1 (5) & 114.86 &   58.49   &   56.21   \\
    Boltz-1 (5) &  173.35  &   37.62   &   31.41   \\
    AF3 (5) & 3,049.41  &   -   &   -   \\
    \bottomrule
  \end{tabular}
\end{table}

As a more formal investigation of the computational resources required to run each baseline method in this work, in Table \ref{table:baseline_method_average_compute_resources} we list the average runtime (in seconds) and peak CPU (GPU) memory usage (in GB) consumed by each method when running them on a 25\% subset of the Astex Diverse dataset. We find that NeuralPLexer provides the lowest computational runtime and DynamicBind the lowest peak CPU and GPU memory requirements during benchmarking.

\section{Documentation for datasets}\label{section:appendix_documentation_for_datasets}

Below, we provide detailed documentation for each dataset included in our benchmark, summarized in Table \ref{table:preprocessed_datasets} of the main text. Each dataset is freely available for download from the benchmark's accompanying Zenodo data record \citep{morehead_2025_19138652} under a CC-BY 4.0 license. In lieu of being able to create associated metadata for each of our macromolecular datasets using an ML-focused library such as Croissant \citep{akhtar2024croissant} (due to file type compatibility \href{https://github.com/mlcommons/croissant/issues/547}{issues}), 
instead, we report structured metadata for our preprocessed datasets using Zenodo's web user interface \citep{morehead_2025_19138652}. Note that, for all datasets, we authors bear all responsibility in case of any violation of rights regarding the usage of such datasets.

\clearpage

\begin{figure}[t]
    \centering
    \begin{subfigure}{0.45\textwidth}
        \centering
        \includegraphics[width=\textwidth]{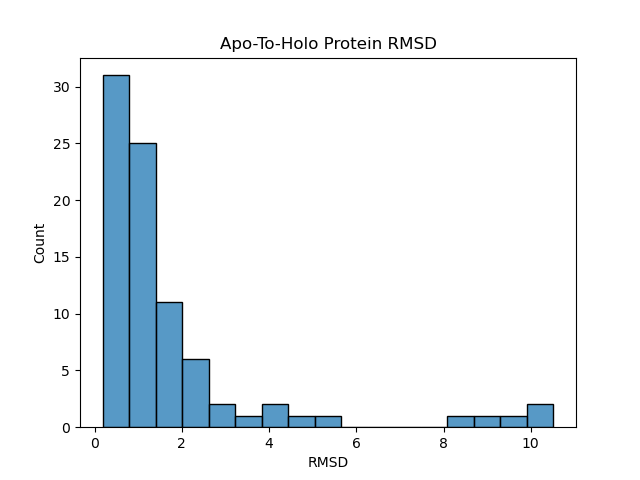}
        \caption{RMSD of AF3's predictions.}
    \end{subfigure}
    \begin{subfigure}{0.45\textwidth}
        \centering
        \includegraphics[width=\textwidth]{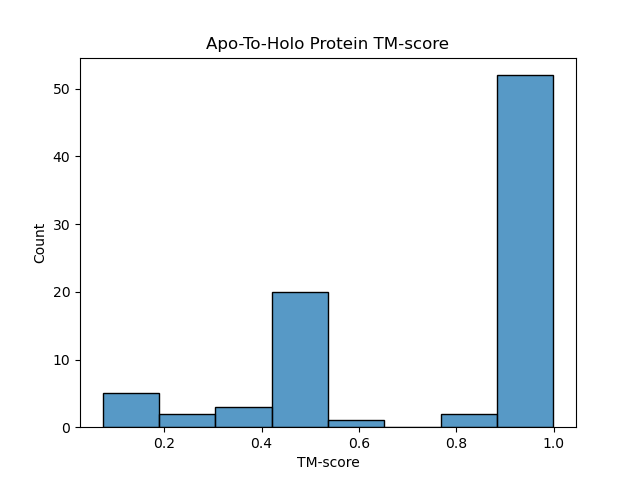}
        \caption{TM-score of AF3's predictions.}
    \end{subfigure}
    \caption{Accuracy of AF3's predicted protein structures for the Astex Diverse dataset.}
    \label{figure:astex_diverse_af3_structure_accuracy}
\end{figure}

\begin{figure}[t]
    \centering
    \begin{subfigure}{0.45\textwidth}
        \centering
        \includegraphics[width=\textwidth]{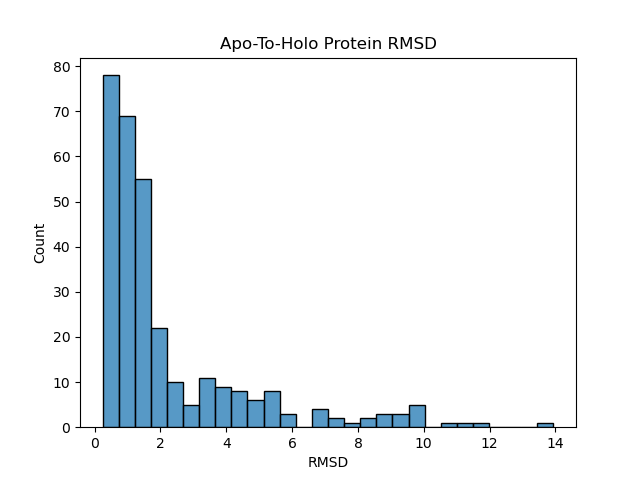}
        \caption{RMSD of AF3's predictions.}
    \end{subfigure}
    \begin{subfigure}{0.45\textwidth}
        \centering
        \includegraphics[width=\textwidth]{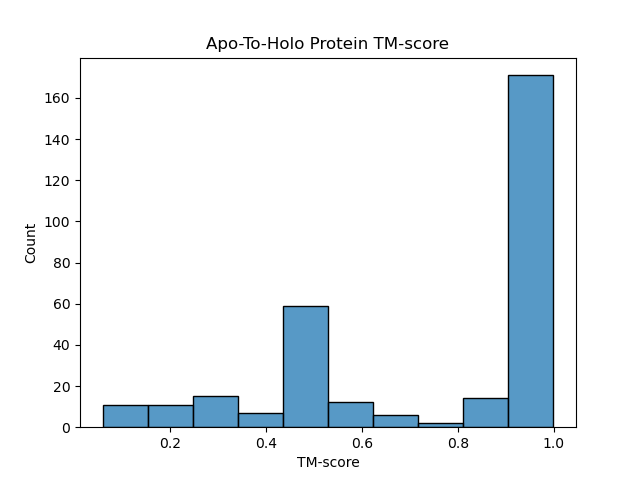}
        \caption{TM-score of AF3's predictions.}
    \end{subfigure}
    \caption{Accuracy of AF3's predicted protein structures for the PoseBusters Benchmark dataset.}
    \label{figure:posebusters_benchmark_af3_structure_accuracy}
\end{figure}

\clearpage

\begin{figure}[t]
    \centering
    \begin{subfigure}{0.45\textwidth}
        \centering
        \includegraphics[width=\textwidth]{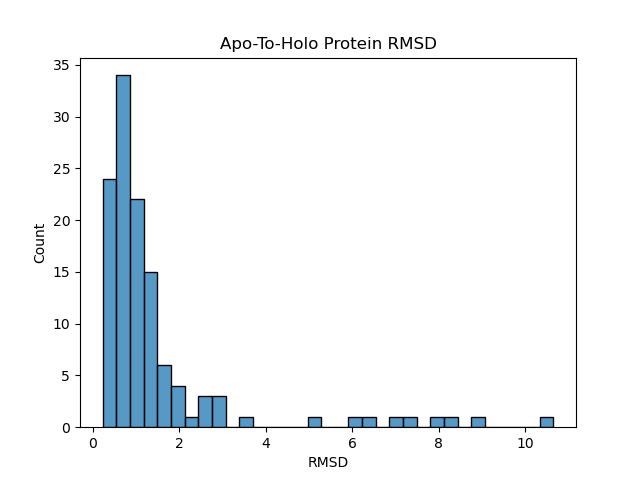}
        \caption{RMSD of AF3's predictions.}
    \end{subfigure}
    \begin{subfigure}{0.45\textwidth}
        \centering
        \includegraphics[width=\textwidth]{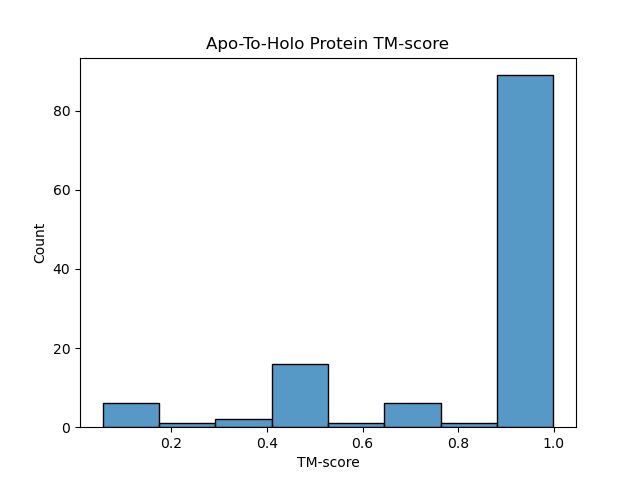}
        \caption{TM-score of AF3's predictions.}
    \end{subfigure}
    \caption{Accuracy of AF3's predicted protein structures for the DockGen dataset.}
    \label{figure:dockgen_af3_structure_accuracy}
\end{figure}

\begin{figure}[t]
    \centering
    \begin{subfigure}{0.45\textwidth}
        \centering
        \includegraphics[width=\textwidth]{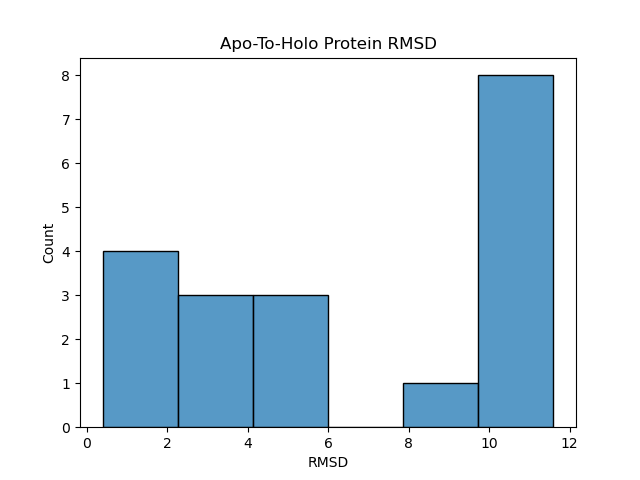}
        \caption{RMSD of AF3's predictions.}
    \end{subfigure}
    \begin{subfigure}{0.45\textwidth}
        \centering
        \includegraphics[width=\textwidth]{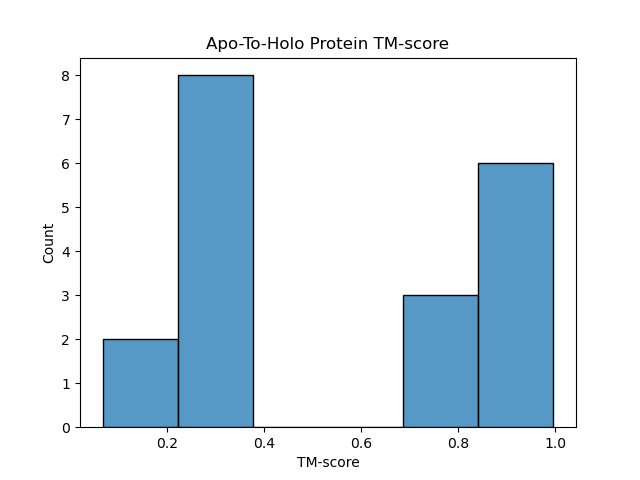}
        \caption{TM-score of AF3's predictions.}
    \end{subfigure}
    \caption{Accuracy of AF3's predicted protein structures for the CASP15 dataset.}
    \label{figure:casp15_af3_structure_accuracy}
\end{figure}

\clearpage

\subsection{Astex Diverse Set - Primary Ligand Docking\\ (Difficulty: \textit{Easy})}\label{section:appendix_astex_diverse_datasheet}

A common drug discovery task is to screen several novel drug-like molecules against a target protein in rapid succession. The Astex Diverse dataset was originally developed with this application in mind, as it features many therapeutically relevant 3D molecules for computational modeling.

\begin{itemize}
    \item \textbf{Motivation} Several downstream drug discovery efforts rely on having access to high-quality molecular data for docking.
    \item \textbf{Collection} For this dataset, which was originally compiled by \citet{hartshorn2007diverse}, we adopt the version further prepared by \citet{buttenschoen2024posebusters}.
    \item \textbf{Composition} The dataset consists of 85 primary ligand protein complexes deposited in the PDB up to 2007. As such, this dataset can be considered an easy benchmarking dataset since many of its complexes may be found in DL methods' PDB-based training datasets. For each of these complexes, we obtained high-accuracy predicted protein structures using AF3. The accuracy of the AF3-predicted structures is measured in terms of their RMSD and TM-score \citep{zhang2004scoring} compared to the corresponding crystal protein structures and is visualized in Figure \ref{figure:astex_diverse_af3_structure_accuracy}. Notably, after alignment with the crystalized (holo) PLI binding pocket residues, 63.53\% (54.12\% with ESMFold) of the predicted structures have a global RMSD below 4 \AA\ and TM-score above 0.7, indicating that most of the dataset's proteins have a reasonably accurate predicted structure.
    \item \textbf{Hosting} Our preprocessed version of the dataset (\url{https://doi.org/10.5281/zenodo.19138652}) can be downloaded from the benchmark's Zenodo data record at \url{https://zenodo.org/records/19138652/files/astex_diverse_set.tar.gz}.
    \item \textbf{Licensing} We have released our preprocessed version of the dataset under a CC-BY 4.0 license. The original PoseBusters Benchmark dataset is available under a CC-BY 4.0 license on Zenodo \citep{buttenschoen_2023_8278563}. The pre-holo-aligned protein structures predicted by AF3 for this dataset
    (available on Zenodo \citep{morehead_2025_19138652})
    must only be used in accordance with AF3's \href{https://github.com/google-deepmind/alphafold3/blob/main/WEIGHTS_TERMS_OF_USE.md}{Terms of Use}.
    \item \textbf{Maintenance} We will announce any errata discovered in or changes made to the dataset using the benchmark's GitHub repository at \url{https://github.com/BioinfoMachineLearning/PoseBench}.
    \item \textbf{Uses} This dataset of predicted (apo) and crystal (holo) protein PDB and crystal (holo) ligand SDF files can be used for primary ligand docking or protein-ligand structure prediction.
    \item \textbf{Metrics} Ligand Centroid RMSD $\leq$ 1 \AA, Ligand Pose RMSD $\leq$ 2 \AA, PoseBusters-Valid (PB-Valid), and PLIF-WM.
\end{itemize}

\clearpage

\subsection{PoseBusters Benchmark Set - Primary Ligand Docking\\ (Difficulty: \textit{Intermediate})}\label{section:appendix_posebusters_benchmark_datasheet}

Like the Astex Diverse dataset, the PoseBusters Benchmark dataset was originally developed for docking individual ligands to target proteins. However, this dataset features a larger and more challenging collection of PLI complexes for computational modeling.

\begin{itemize}
    \item \textbf{Motivation} Data sources of challenging primary ligand protein complexes for molecular docking are critical for the development of future docking methods.
    \item \textbf{Collection} For this dataset, we adopt the version introduced by \citet{buttenschoen2024posebusters}.
    \item \textbf{Composition} The dataset consists of 308 primary ligand protein complexes deposited in the PDB in 2019 and after. As such, this dataset poses a moderate challenge for DL methods, since several of such methods were trained on data deposited before this cutoff date (notably except for Chai-1 and AF3/Boltz-1 which used training cutoff dates of January 12, 2021 and September 30, 2021, respectively). For each of the dataset's complexes, we obtained high-accuracy predicted protein structures using AF3. The accuracy of the AF3-predicted structures is measured in terms of their RMSD and TM-score compared to the corresponding crystal protein structures and is visualized in Figure \ref{figure:posebusters_benchmark_af3_structure_accuracy}. Notably, after alignment with the crystalized (holo) PLI binding pocket residues, 59.09\% (53.25\% with ESMFold) of the predicted structures have a global RMSD below 4 \AA\ and TM-score above 0.7, indicating that most of the dataset's proteins have a reasonably accurate predicted structure.
    \item \textbf{Hosting} Our preprocessed version of the dataset (\url{https://doi.org/10.5281/zenodo.19138652}) can be downloaded from the benchmark's Zenodo data record at \url{https://zenodo.org/records/19138652/files/posebusters_benchmark_set.tar.gz}.
    \item \textbf{Licensing} We have released our preprocessed version of the dataset under a CC-BY 4.0 license. The original dataset is available under a CC-BY 4.0 license on Zenodo \citep{buttenschoen_2023_8278563}. The pre-holo-aligned protein structures predicted by AF3 for this dataset (available on Zenodo \citep{morehead_2025_19138652}) must only be used in accordance with AF3's \href{https://github.com/google-deepmind/alphafold3/blob/main/WEIGHTS_TERMS_OF_USE.md}{Terms of Use}.
    \item \textbf{Maintenance} We will announce any errata discovered in or changes made to the dataset using the benchmark's GitHub repository at \url{https://github.com/BioinfoMachineLearning/PoseBench}.
    \item \textbf{Uses} This dataset of predicted (apo) and crystal (holo) protein PDB and crystal (holo) ligand SDF files can be used for primary ligand docking or protein-ligand structure prediction.
    \item \textbf{Metrics} Ligand Centroid RMSD $\leq$ 1 \AA, Ligand Pose RMSD $\leq$ 2 \AA, PoseBusters-Valid (PB-Valid), and PLIF-WM.
\end{itemize}

\clearpage

\subsection{DockGen-E Set - Primary Ligand Docking\\ (Difficulty: \textit{Challenging})}\label{section:appendix_dockgen_datasheet}

The DockGen dataset was originally designed for binding individual ligands to target proteins within functionally novel PLI binding pockets \citep{corso2024deep}, filtering out any protein chains not associated with a novel pocket, which can remove important biomolecular context for DL methods to make their predictions. In this work, we introduced DockGen-E, an enhanced version of DockGen that has each method predict the full biologically relevant assembly of each novel pocket to expand their structural prediction contexts (n.b., which is specifically important to achieve best performance with DL co-folding methods such as AF3). As such, this new dataset is useful for evaluating how well each baseline method can predict complexes containing functionally distinct binding pockets compared to those on which the method may have \textit{primarily} been trained.

\begin{itemize}
    \item \textbf{Motivation} Data sources of PLI complexes representing novel primary ligand binding pockets are critical for the development of generalizable docking methods.
    \item \textbf{Collection} To curate this dataset, we collected the original dataset's protein and ligand binding pocket annotations for each complex introduced by \citet{corso2024deep}. Subsequently, we retrieved the corresponding first biological assembly listed in the PDB to obtain each PDB entry's biologically relevant complex, filtering out complexes for which the first assembly could not be mapped to its original protein and ligand binding pocket annotation. This procedure left 122 biologically relevant assemblies remaining for benchmarking. Important to note is that these original DockGen complexes were deposited in the PDB from 2019 onward, making this benchmarking dataset partially overlap with the training datasets of multiple DL co-folding baseline methods such as NeuralPLexer, Chai-1, Boltz-1, and AF3. Nonetheless, our benchmarking results in the main text demonstrate that baseline DL methods are challenged to find the correct (novel) binding pocket conformations represented by this dataset, suggesting that all baseline DL models have yet to learn truly comprehensive representations of protein-ligand binding.
    \item \textbf{Composition} The dataset consists of 122 primary ligand protein complexes, for each of which we obtained high-accuracy predicted protein structures using AF3. The accuracy of the AF3-predicted structures is measured in terms of their RMSD and TM-score compared to the corresponding crystal protein structures and is visualized in Figure \ref{figure:dockgen_af3_structure_accuracy}. Notably, after alignment with the crystalized (holo) PLI binding pocket residues, 74.59\% (57.38\% with ESMFold) of the predicted structures have a global RMSD below 4 \AA\ and TM-score above 0.7, indicating that most of the dataset's proteins have a reasonably accurate predicted structure.
    \item \textbf{Hosting} Our preprocessed version of the dataset (\url{https://doi.org/10.5281/zenodo.19138652}) can be downloaded from the benchmark's Zenodo data record at \url{https://zenodo.org/records/19138652/files/dockgen_set.tar.gz}.
    \item \textbf{Licensing} We have released our preprocessed version of the DockGen-E dataset under a CC-BY 4.0 license. The original DockGen dataset is available under an MIT license on Zenodo \citep{corso_2024_10656052}, and the DockGen-E dataset along with its pre-holo-aligned protein structures predicted by AF3 is also available on Zenodo \citep{morehead_2025_19138652}. Notably, these AF3-predicted protein structures must only be used in accordance with AF3's \href{https://github.com/google-deepmind/alphafold3/blob/main/WEIGHTS_TERMS_OF_USE.md}{Terms of Use}.
    \item \textbf{Maintenance} We will announce any errata discovered in or changes made to the dataset using the benchmark's GitHub repository at \url{https://github.com/BioinfoMachineLearning/PoseBench}.
    \item \textbf{Uses} This dataset of predicted (apo) and crystal (holo) protein PDB and crystal (holo) ligand PDB files can be used for primary ligand docking or protein-ligand structure prediction.
    \item \textbf{Metrics} Ligand Centroid RMSD $\leq$ 1 \AA, Ligand Pose RMSD $\leq$ 2 \AA, PoseBusters-Valid (PB-Valid), and PLIF-WM.
\end{itemize}

\clearpage

\subsection{CASP15 Set - Multi-Ligand Docking\\ (Difficulty: \textit{Challenging})}\label{section:appendix_casp15_datasheet}

As the most distinct of our benchmark's four evaluation datasets, the CASP15 PLI dataset was created to represent the new protein-ligand modeling category in the 15th Critical Assessment of Techniques for Structure Prediction (CASP) competition. Whereas datasets such as PoseBusters Benchmark and Astex Diverse feature solely primary ligand protein complexes, the CASP15 dataset provides the research community with a variety of challenging organic (e.g., drug molecules) and inorganic (e..g., ion) cofactors for \textit{multi}-ligand biomolecular modeling and scoring.

\begin{itemize}
    \item \textbf{Motivation} Multi-ligand evaluation datasets for molecular docking provide the rare opportunity to assess how well baseline methods can model intricate PLIs while avoiding troublesome inter-ligand steric clashes. Additionally, accurate modeling of multi-ligand complexes in future work may lead to improved algorithms for computational enzyme design and regulation \citep{stark2023harmonic}.
    \item \textbf{Collection} For this dataset, we manually collect each publicly and privately available CASP15 protein-bound ligand complex structure compatible with protein-ligand (e.g., non-nucleic acid) benchmarking.
    \item \textbf{Composition} The dataset consists of 102 (86) fragment ligands contained within 19 (15) separate (publicly available) protein complexes, of which 6 (2) and 13 (2) of these complexes are single and multi-ligand complexes, respectively. Importantly, each of such complexes (if publicly available) was released in the PDB after 2022, making this benchmarking dataset strictly non-overlapping with the training datasets of all baseline methods. The accuracy of the dataset's AF3-predicted structures is measured in terms of their RMSD and TM-score compared to the corresponding crystal protein structures and is visualized in Figure \ref{figure:casp15_af3_structure_accuracy}. Notably, after alignment with the crystalized (holo) PLI binding pocket residues, 36.84\% and 20.00\% (26.32\% and 13.33\% with ESMFold) of the total and publicly available predicted structures, respectively, have a global RMSD below 4 \AA\ and TM-score above 0.7, indicating that a portion of the dataset's proteins has a reasonably accurate predicted structure. Given the much larger structural assemblies of this dataset's protein complexes compared to those of the other benchmark datasets, we believe the accuracy of these predictions may be improved with advancements in machine learning modeling of biomolecular assemblies.
    \item \textbf{Hosting} Our preprocessed version of (the publicly available version of) this dataset (\url{https://doi.org/10.5281/zenodo.19138652}) can be downloaded from the benchmark's Zenodo data record at \url{https://zenodo.org/records/19138652/files/casp15_set.tar.gz}.
    \item \textbf{Licensing} We have released our preprocessed version of the (public) dataset under a CC-BY 4.0 license. The original (public) dataset is free for download via the RCSB PDB \citep{bank1971protein}, whereas the dataset's remaining (private) complexes must be manually requested from the CASP organizers. The pre-holo-aligned protein structures predicted by AF3 for this dataset (available on Zenodo \citep{morehead_2025_19138652}) must only be used in accordance with AF3's \href{https://github.com/google-deepmind/alphafold3/blob/main/WEIGHTS_TERMS_OF_USE.md}{Terms of Use}.
    \item \textbf{Maintenance} We will announce any errata discovered in or changes made to the dataset using the benchmark's GitHub repository at \url{https://github.com/BioinfoMachineLearning/PoseBench}.
    \item \textbf{Uses} This dataset of predicted (apo) and crystal (holo) protein PDB and crystal (holo) ligand PDB files can be used for multi-ligand docking or protein-ligand structure prediction.
    \item \textbf{Metrics} (Fragment) Ligand Pose RMSD $\leq$ 2 \AA, (Complex) PoseBusters-Valid (PB-Valid), and (Complex) PLIF-WM.
\end{itemize}

\clearpage

\section{Analysis of protein-ligand interactions}
\label{section:appendix_analysis_of_protein_ligand_interactions}

\subsection{Dataset protein-ligand interaction distributions}
\label{section:appendix_analysis_of_dataset_protein_ligand_interactions}

\begin{figure}
  \centering
  \includegraphics[width=\linewidth]{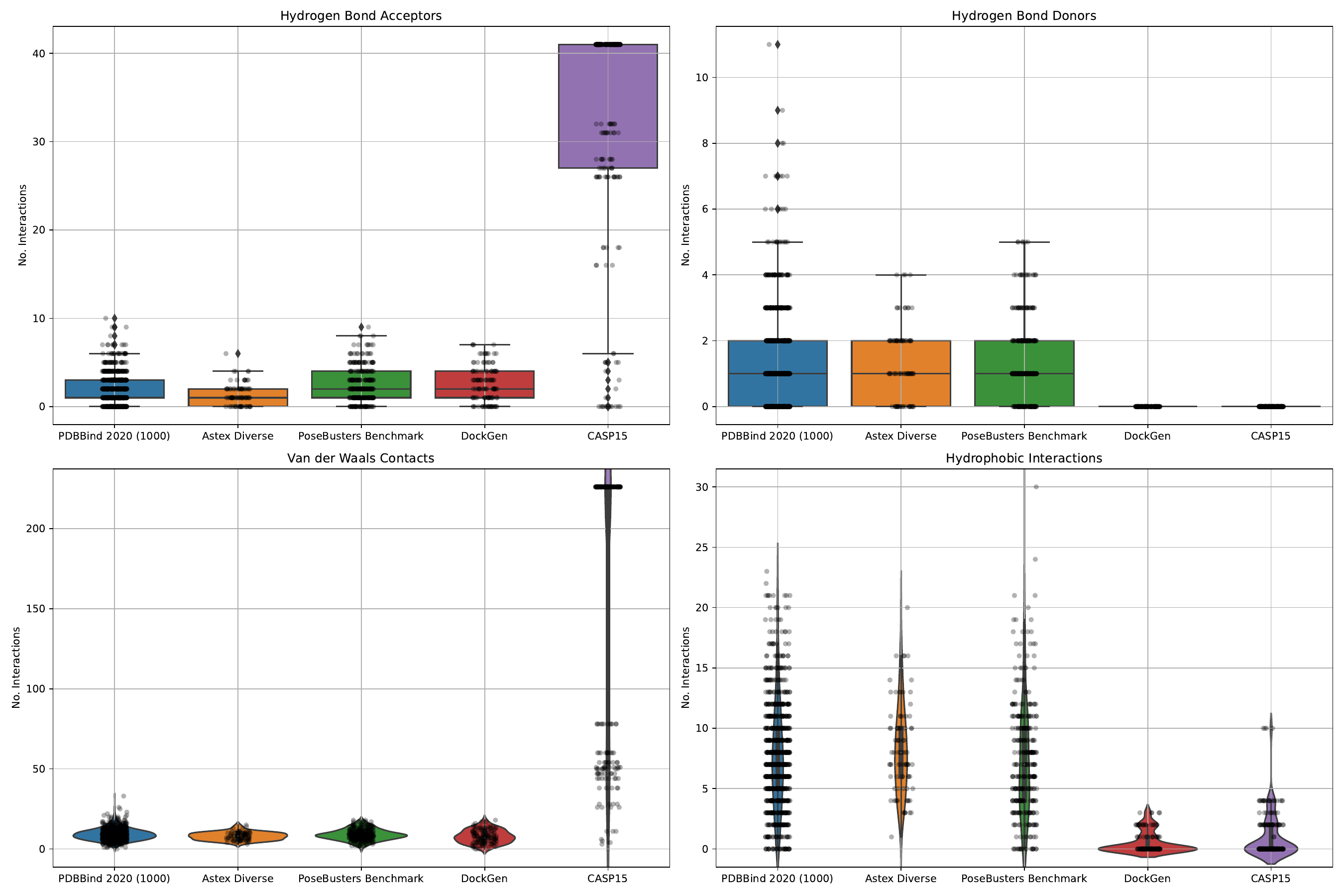}
  \caption{Comparative analysis of evaluation dataset protein-ligand interactions.}
  \label{figure:dataset_interaction_analysis}
\end{figure}

Inspired by a similar analysis presented in the PoseCheck benchmark \citep{harris2023benchmarking}, in this section, we study the frequency of different types of protein-ligand (pocket-level) interactions such as van der Waals contacts and hydrophobic interactions occurring natively within (n.b., a size-1000 random subset of) the commonly-used PDBBind 2020 docking training dataset (i.e., PDBBind 2020 (1000)) as well as the Astex Diverse, PoseBusters Benchmark, DockGen, and CASP15 benchmark datasets, respectively. In particular, these measures allow us to better understand the diversity of interactions each baseline method within the \textsc{PoseBench} benchmark is tasked to model, within the context of each evaluation dataset. Furthermore, these measures directly indicate which benchmark datasets are most \textit{dissimilar} from commonly used training data for baseline methods. Figure \ref{figure:dataset_interaction_analysis} displays the results of this analysis.

Overall, we find that the PDBBind 2020, Astex Diverse, and PoseBusters Benchmark datasets contain similar types and frequencies of interactions, with the PoseBusters Benchmark dataset containing slightly more hydrogen bond acceptors ($\sim$3 vs 1) and fewer van der Waals contacts ($\sim$5 vs 8) on average compared to the PDBBind 2020 dataset. However, we observe a more notable difference in interaction types and frequencies between the DockGen and CASP15 datasets and the three other datasets. Specifically, we find these two benchmark datasets contain a notably different quantity of hydrogen bond acceptors and donors (n.b., $\sim$40 for CASP15), van der Waals contacts ($\sim$200 for CASP15), and hydrophobic interactions ($\sim$2 for DockGen) on average. These dataset-level interaction disparities may partially explain the baseline-challenging DockGen benchmarking results reported in Section \ref{section:results_and_discussion} of the main text.

Also particularly interesting to note is the CASP15 dataset's bimodal distribution of van der Waals contacts, suggesting that the dataset contains two primary classes of interacting ligands giving rise to van der Waals interactions. One possible explanation for this phenomenon is that the CASP15 prediction targets, in contrast to the PDBBind, Astex Diverse, PoseBusters Benchmark, and DockGen targets, consist of a variety of both organic (e.g., drug-like molecules) and inorganic (e.g., metal) cofactors.

\subsection{Baseline method protein-ligand interaction distributions}
\label{section:appendix_analysis_of_method_protein_ligand_interactions}

\begin{figure}
  \centering
  \includegraphics[width=\linewidth]{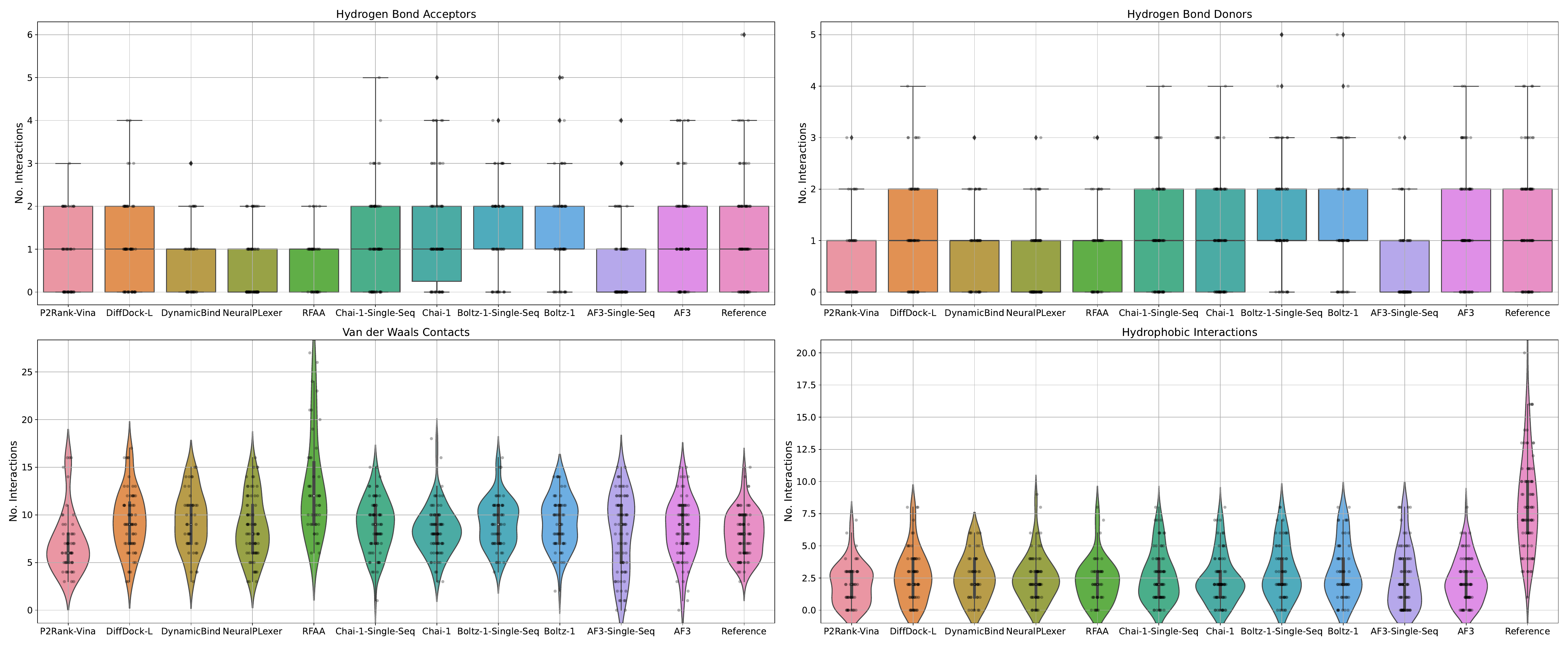}
  \caption{Comparative analysis of Astex Diverse protein-ligand interactions.}
  \label{figure:astex_method_interaction_analysis}
\end{figure}

\begin{figure}
  \centering
  \includegraphics[width=\linewidth]{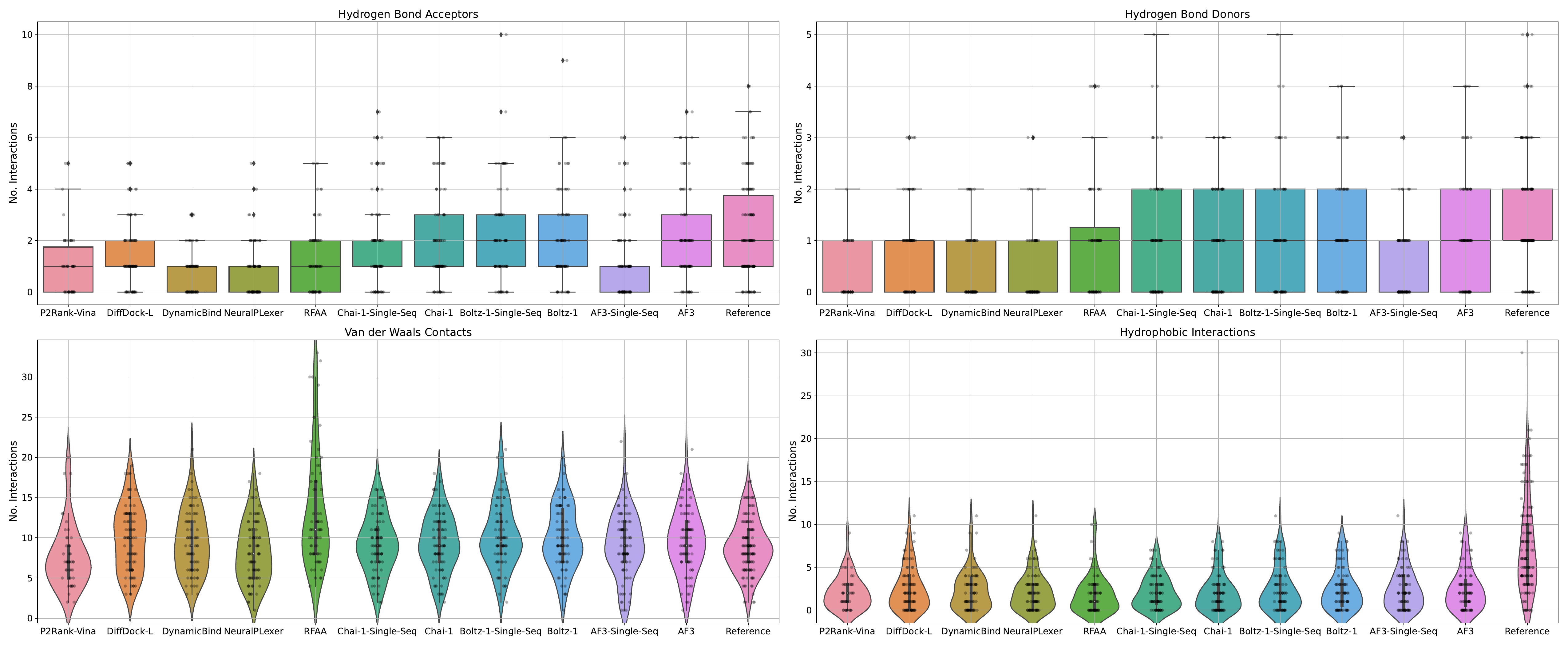}
  \caption{Comparative analysis of PoseBusters Benchmark protein-ligand interactions.}
  \label{figure:posebusters_method_interaction_analysis}
\end{figure}

\begin{figure}
  \centering
  \includegraphics[width=\linewidth]{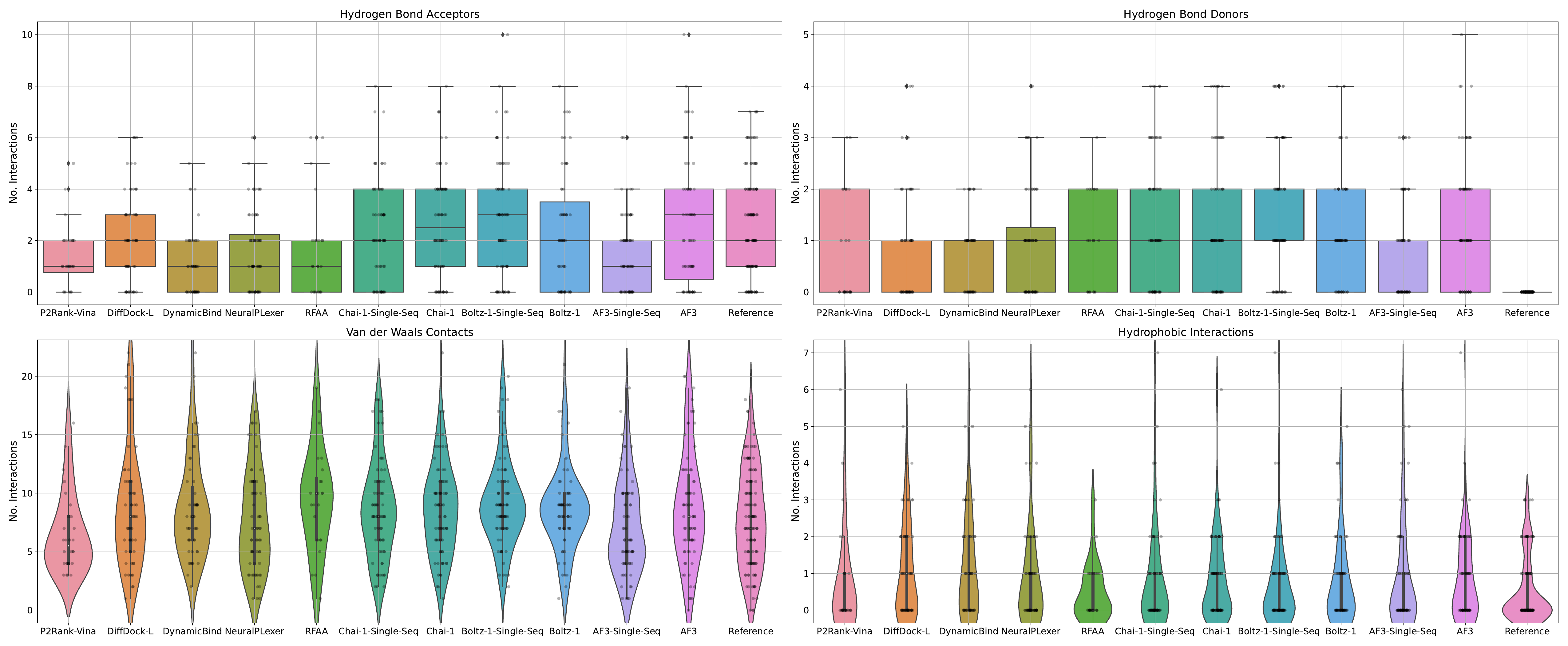}
  \caption{Comparative analysis of DockGen protein-ligand interactions.}
  \label{figure:dockgen_method_interaction_analysis}
\end{figure}

\begin{figure}
  \centering
  \includegraphics[width=\linewidth]{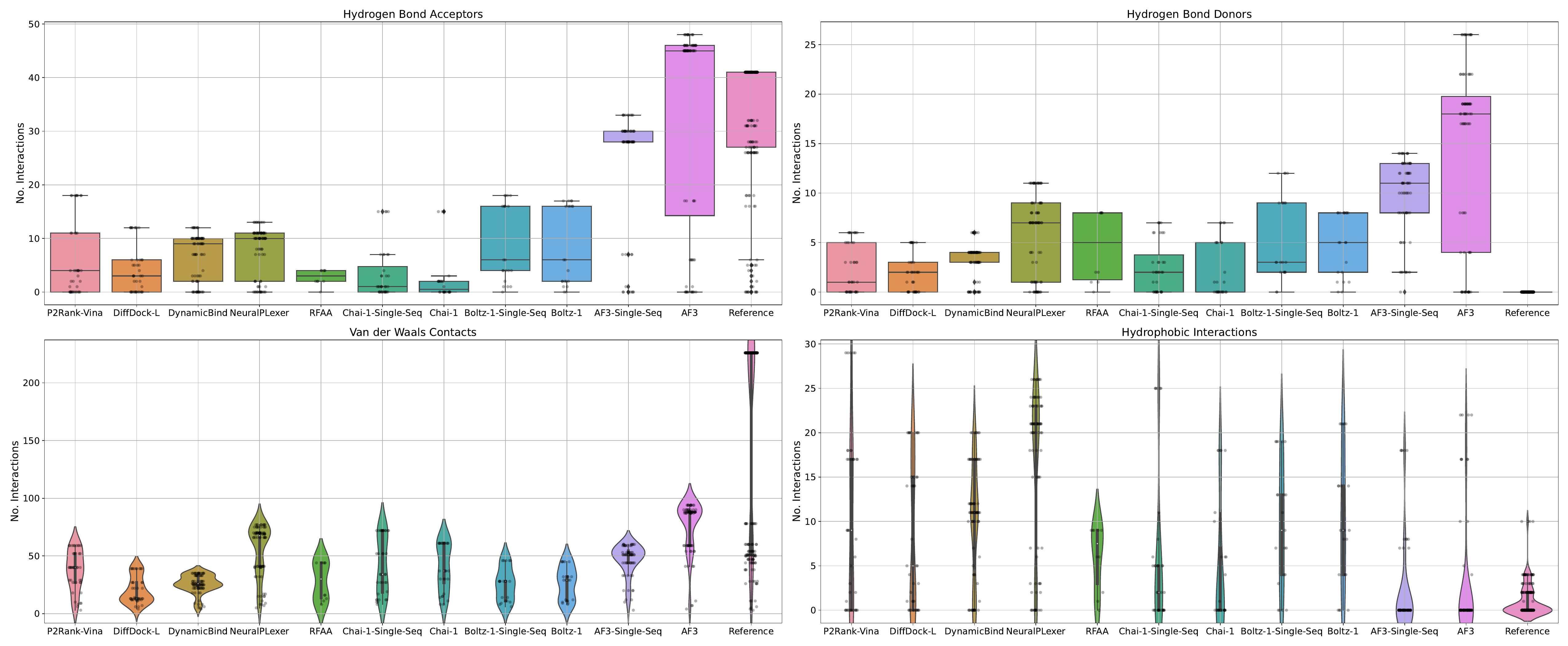}
  \caption{Comparative analysis of CASP15 protein-ligand interactions.}
  \label{figure:casp15_method_interaction_analysis}
\end{figure}

Intrigued by the dataset interaction patterns presented in Figure \ref{figure:dataset_interaction_analysis}, here we further investigate the predicted PLIs produced by each baseline method for each evaluation dataset to study which DL methods can most faithfully reproduce the native distribution of PLIs within each dataset. Our results in Figures \ref{figure:astex_method_interaction_analysis}, \ref{figure:posebusters_method_interaction_analysis}, \ref{figure:dockgen_method_interaction_analysis}, and \ref{figure:casp15_method_interaction_analysis} suggest that AF3 demonstrates the best overall ability to recapitulate the crystalized PLIs observed within these datasets, in line with the PLIF-WM benchmarking results presented in Section \ref{section:results_and_discussion} of the main text. Nonetheless, its predicted interaction distributions, in particular for the DockGen and CASP15 datasets, have much room for improvement, especially for more structured interactions such as hydrogen bonds.

\clearpage

\clearpage

\clearpage

\section{Additional method descriptions}\label{section:appendix_additional_method_descriptions}

To better contextualize the benchmark's results comparing DL methods to conventional docking algorithms, in this section, we provide further details regarding how each baseline method in the benchmark leverages different sources of biomolecular information to predict PLIs for a given protein target.

\subsection{Input and output formats}\label{section:appendix_input_and_output_formats}

\begin{enumerate}[label=\arabic*., ref=\arabic*]
  \item Formats for \underline{conventional} methods are as follows:
  \begin{enumerate}[label=\alph*), ref=\theenumi\alph*]
    \item Molecular docking (protein-fixed) software tools such as \textbf{AutoDock Vina}, which require specification of protein binding sites, are provided with not only a predicted protein structure from AF3 but also the centroid coordinates of each predicted PLI binding site residue as estimated by the well-known P2Rank binding site prediction algorithm \citep{krivak2018p2rank}. Such binding site residues are classified using a 10 \AA\ protein-ligand heavy atom interaction threshold and a 25 \AA\ inter-ligand heavy atom interaction threshold to define a "group" of ligands belonging to the same binding site and therefore residing in the same 25 \AA$^3$-sized binding site input voxel for AutoDock Vina. For interested readers, for all four benchmark datasets, we also provide the benchmarking code necessary to run AutoDock Vina using any other baseline method's predicted binding site residues (e.g., those of DiffDock-L) according to the same binding site classification scheme described above.
  \end{enumerate}

  \item Formats for \underline{DL docking} methods are as follows:
  \begin{enumerate}[label=\alph*), ref=\theenumi\alph*]
    \item \textbf{DiffDock-L} is provided with a protein structure predicted by AF3 and (fragment) ligand SMILES strings. The model is then tasked with producing (multiple rank-ordered) ligand conformations (for each fragment) for the given protein structure (which remains fixed during docking). Note that DiffDock-L does not natively support multi-ligand SMILES string inputs, so in this work, we propose a modified inference procedure for DiffDock-L which \textit{autoregressively} presents each (fragment) ligand SMILES string to the model while providing the same predicted protein structure to the model in each inference iteration (reporting for each complex the average confidence score over all iterations). Notably, as an inference-time modification, this sampling formulation permits multi-ligand sampling yet cannot model multi-ligand interactions directly and therefore often produces inter-ligand steric clashes.
    
    \item As a single-ligand DL (flexible) docking method, \textbf{DynamicBind} adopts the same input and output formats as DiffDock-L with the following exceptions: (1) the predicted input protein structure is now flexible in response to (fragment) ligand docking; (2) the autoregressive inference procedure we adapted from that of DiffDock-L now provides DynamicBind with its own most recently predicted protein structure in each (fragment) ligand inference iteration, thereby providing the model with partial multi-ligand interaction context; and (3) iteration-averaged confidence scores \textit{and} predicted affinities are reported for each complex. Nonetheless, for both DiffDock-L and DynamicBind, such modified inference procedures highlight the importance in future work of retraining such generative docking methods directly on multi-ligand complexes to address such inference-time compromises.
  \end{enumerate}

  \item Formats for \underline{DL co-folding} methods are as follows:
  \begin{enumerate}[label=\alph*), ref=\theenumi\alph*]
    \item One of the first DL co-folding methods, \textbf{RoseTTAFold-All-Atom} is provided with a (multi-chain) protein sequence as well as (fragment) ligand SMILES strings. The method is subsequently tasked with producing not only a (single) bound ligand conformation but also the bound protein conformation, using diverse MSA databases to provide evolutionary information to the model.

    \item \textbf{NeuralPLexer} is a protein-ligand co-folding diffusion model trained using expansive PDB molecule and protein data sources. It receives as its inputs a (multi-chain) protein sequence as well as (fragment) ligand SMILES strings. The method is then tasked with producing multiple rank-ordered (flexible) protein-ligand structure conformations for each input complex, where we use the method's average ligand heavy atom plDDT scores for sampling ranking.

    \item \textbf{AlphaFold 3} is a commercially-restricted biomolecular co-folding model trained on exhaustive PDB crystal structures and AlphaFold 2-predicted distillation structures. Following its default settings for inference, the model receives as its inputs a (multi-chain) protein sequence and (fragment) ligand SMILES strings, with default MSA and template inputs provided to the model. The method is then tasked with producing multiple rank-ordered (flexible) protein-ligand structure conformations for each input complex, using the method's intrinsic ranking score \citep{abramson2024accurate} for rank-ordering.

    \item \textbf{Chai-1} is an open-source co-folding model (akin to AF3) trained on exhaustive PDB crystal structures and AlphaFold 2-predicted distillation structures along with AF3-based training protocols. Following its default settings for inference, the model receives as its inputs a (multi-chain) protein sequence and (fragment) ligand SMILES strings, with paired MSAs yet no template structures provided (as is its default setting). The method is then tasked with producing multiple rank-ordered protein-ligand bound structure conformations for each input complex, using the method's intrinsic AF3-like ranking score for rank-ordering. Note that, as Chai-1's source code does not provide resources to generate multiple sequence alignments for input featurization, Chai-1 uses standardized (taxonomy-paired) multiple sequence alignments akin to those used by AF3 in all benchmarking experiments.

    \item \textbf{Boltz-1} is an open-source co-folding model (akin to AF3 and Chai-1) trained on exhaustive PDB crystal structures and AlphaFold 2-predicted distillation structures along with AF3-based training protocols. By default, the model receives as its inputs a (multi-chain) protein sequence and (fragment) ligand SMILES strings, with paired MSAs yet no template structures provided (as is its default setting). The method then produces multiple rank-ordered protein-ligand bound structure conformations for each input complex, using the method's intrinsic AF3-like ranking quantity for rank-ordering. Note that, for the sake of benchmarking consistency (n.b., versus Chai-1), Boltz-1 uses standardized (taxonomy-paired) multiple sequence alignments akin to those used by AF3 in all benchmarking experiments. Note that, by default, we evaluate Boltz-1 with its inference-time potential functions enabled (i.e., Boltz-1x), making its predictions slightly slower but considerably more physically plausible overall (according to the PoseBusters software suite \citep{buttenschoen_2023_8278563}), and we do not evaluate Boltz-2 (trained on PDB data deposited up to June 1, 2023) to maintain the validity of our benchmark's time-splits.
  \end{enumerate}
  
\end{enumerate}

\clearpage

\section{Additional results}
\label{section:appendix_additional_results}

In this section, we provide additional results for each baseline method using the Astex Diverse, PoseBusters Benchmark, and DockGen datasets as well as the CASP15 ligand prediction targets. Note that for all violin plots listed in this section, we curate them using combined results across each method's three independent runs (where applicable), in contrast to this section's bar charts where we instead report mean and standard deviation values across each method's three independent runs.

\subsection{Expanded primary ligand results}
\label{section:appendix_expanded_primary_ligand_results}

\subsubsection{Primary ligand RMSD results}
\label{section:appendix_primary_ligand_rmsd_results}

\begin{figure}
  \centering
  \includegraphics[width=\linewidth]{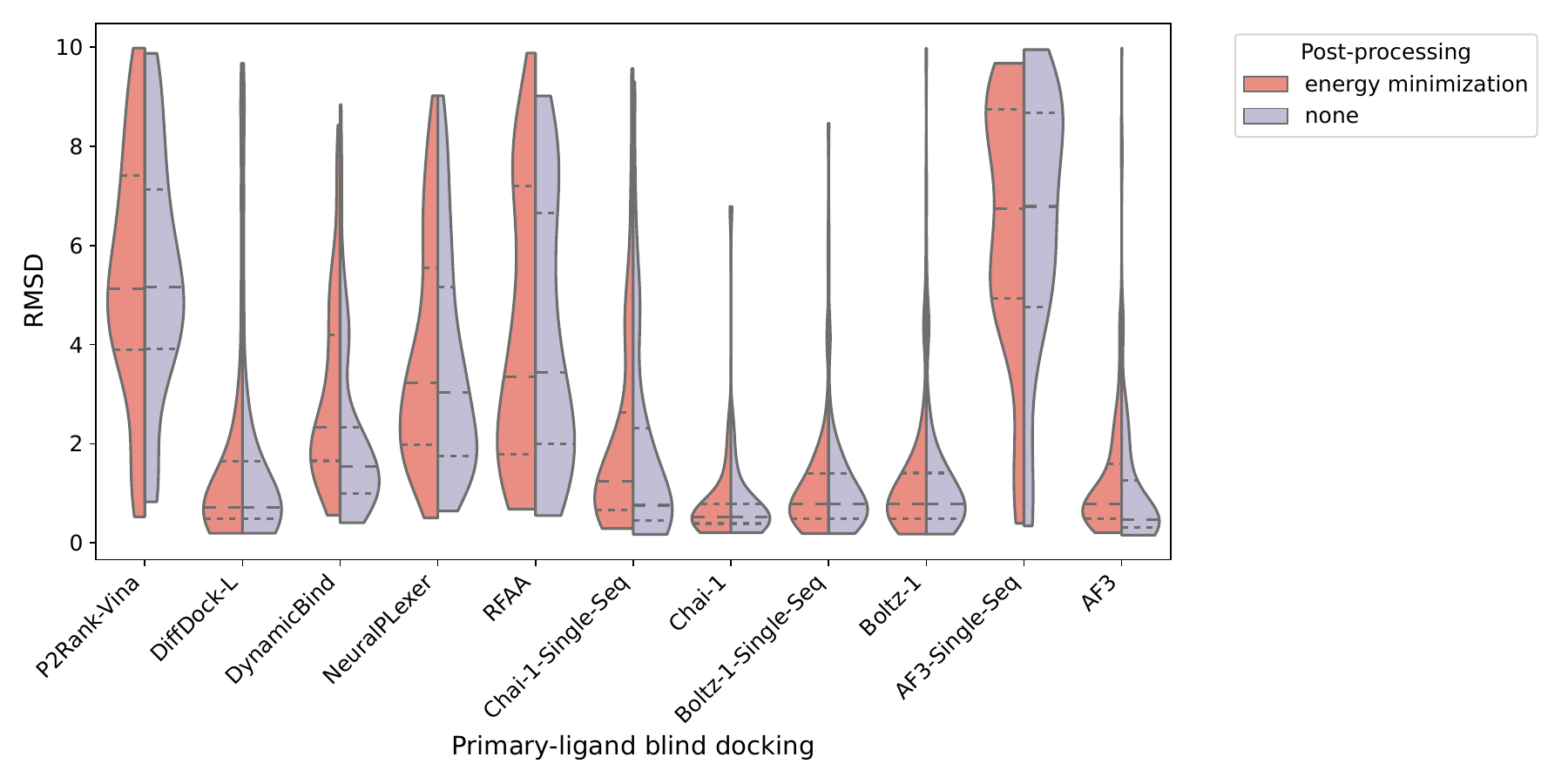}
  \caption{Astex Diverse dataset results for primary ligand docking RMSD.}
  \label{figure:astex_diverse_primary_ligand_relaxed_rmsd_violin_plot}
\end{figure}

\begin{figure}
  \centering
  \includegraphics[width=\linewidth]{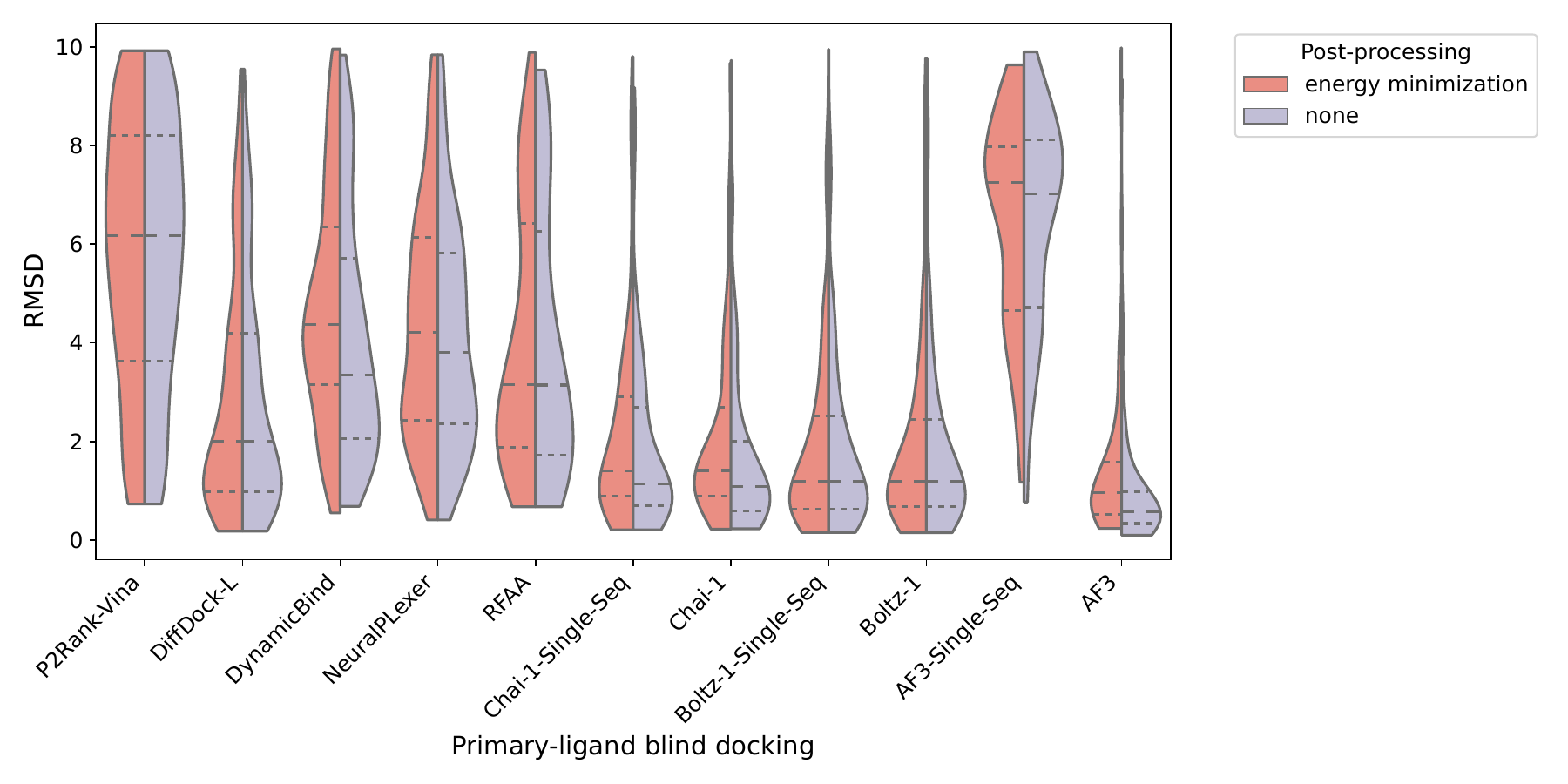}
  \caption{PoseBusters Benchmark dataset results for primary ligand docking RMSD.}
  \label{figure:posebusters_benchmark_primary_ligand_relaxed_rmsd_violin_plot}
\end{figure}

\begin{figure}
  \centering
  \includegraphics[width=\linewidth]{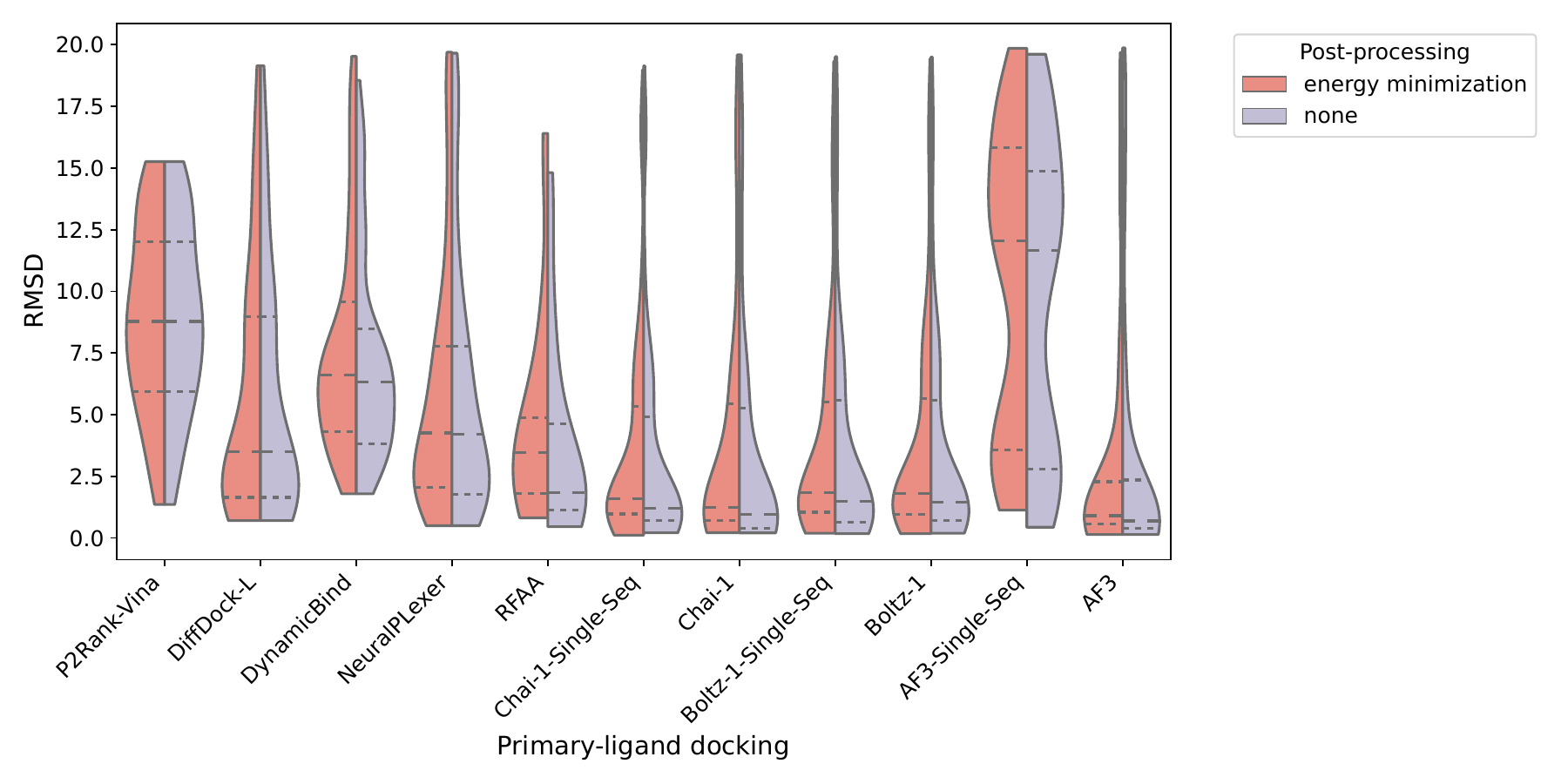}
  \caption{DockGen dataset results for primary ligand docking RMSD.}
  \label{figure:dockgen_primary_ligand_relaxed_rmsd_violin_plot}
\end{figure}

In Figures \ref{figure:astex_diverse_primary_ligand_relaxed_rmsd_violin_plot}, \ref{figure:posebusters_benchmark_primary_ligand_relaxed_rmsd_violin_plot}, and \ref{figure:dockgen_primary_ligand_relaxed_rmsd_violin_plot}, we report the (binding site-superimposed) ligand RMSD values of each baseline method across the primary ligand Astex Diverse, PoseBusters Benchmark, and DockGen datasets, with molecular dynamics (MD)-based structural relaxation applied post-hoc. Overall, these figures demonstrate that Chai-1, Boltz-1, and AF3 achieve the tightest RMSD distributions, except for single-sequence AF3 which frequently produces catastrophic prediction errors by targeting incorrect PLI binding pockets. Further, these results show that MD-based relaxation generally does not modify the RMSD distribution of most baseline methods, except for RFAA and AF3 for which neither seem to benefit from such post-hoc optimizations.

\clearpage

\clearpage

\subsection{Expanded CASP15 results}
\label{section:appendix_additional_casp15_results}

\subsubsection{Overview of expanded results}
In this section, we begin by reporting additional CASP15 benchmarking results in terms of each baseline method's multi-ligand RMSD and lDDT-PLI distributions as violin plots. Subsequently, we report successful ligand docking success rates as well as RMSD and lDDT-PLI results specifically for the single-ligand (i.e., primary ligand) CASP15 targets. Lastly, we report all the above single and multi-ligand results specifically using only the CASP15 targets for which the crystal structures are publicly available, to facilitate reproducible future benchmarking.

\subsubsection{Multi-ligand RMSD and lDDT-PLI}
\begin{figure}
  \centering
  \includegraphics[width=\linewidth]{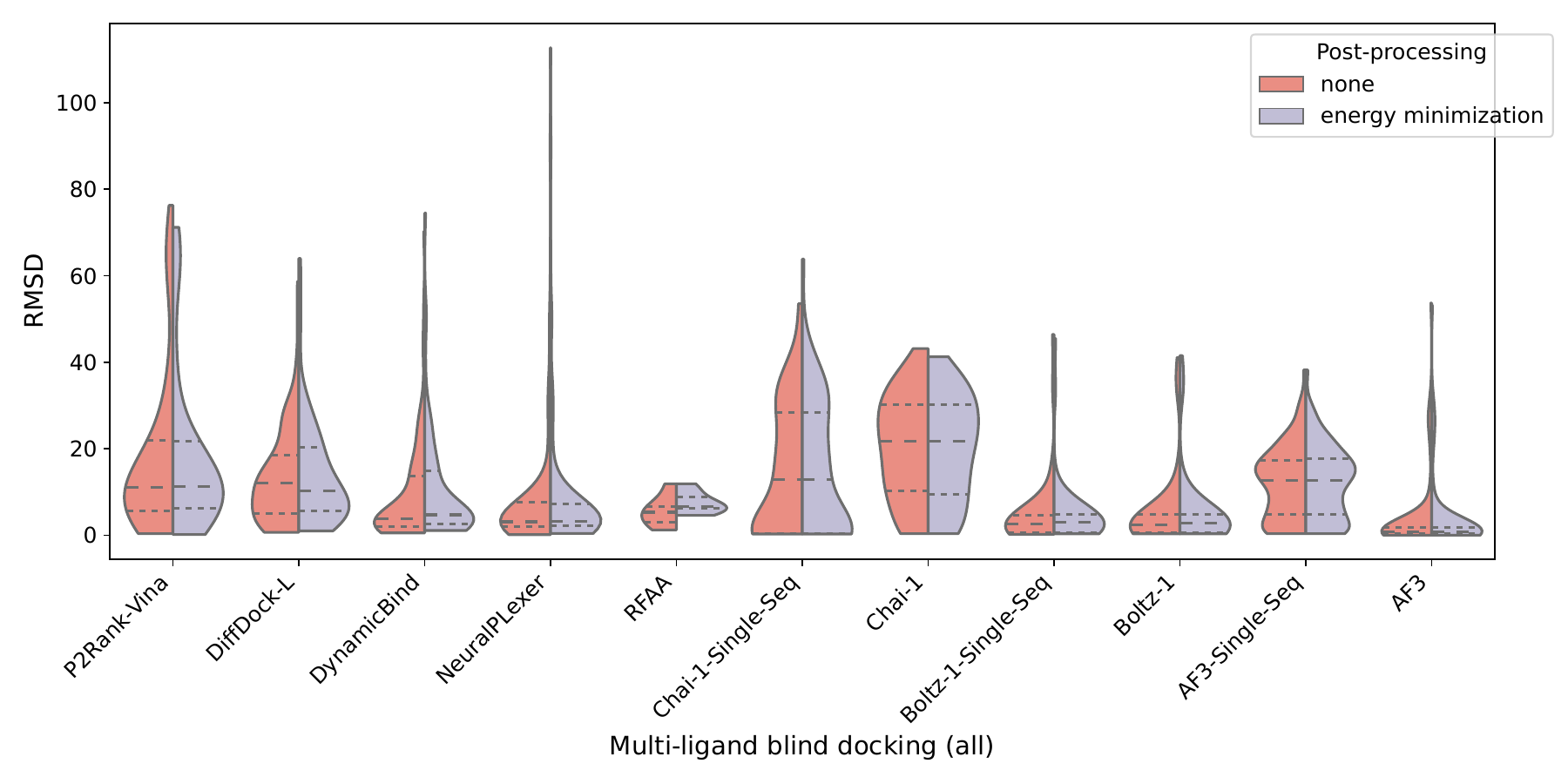}
  \caption{CASP15 dataset results for multi-ligand docking RMSD with relaxation.}
  \label{figure:casp15_multi_ligand_rmsd_results_with_relaxation}
\end{figure}

\begin{figure}
  \centering
  \includegraphics[width=\linewidth]{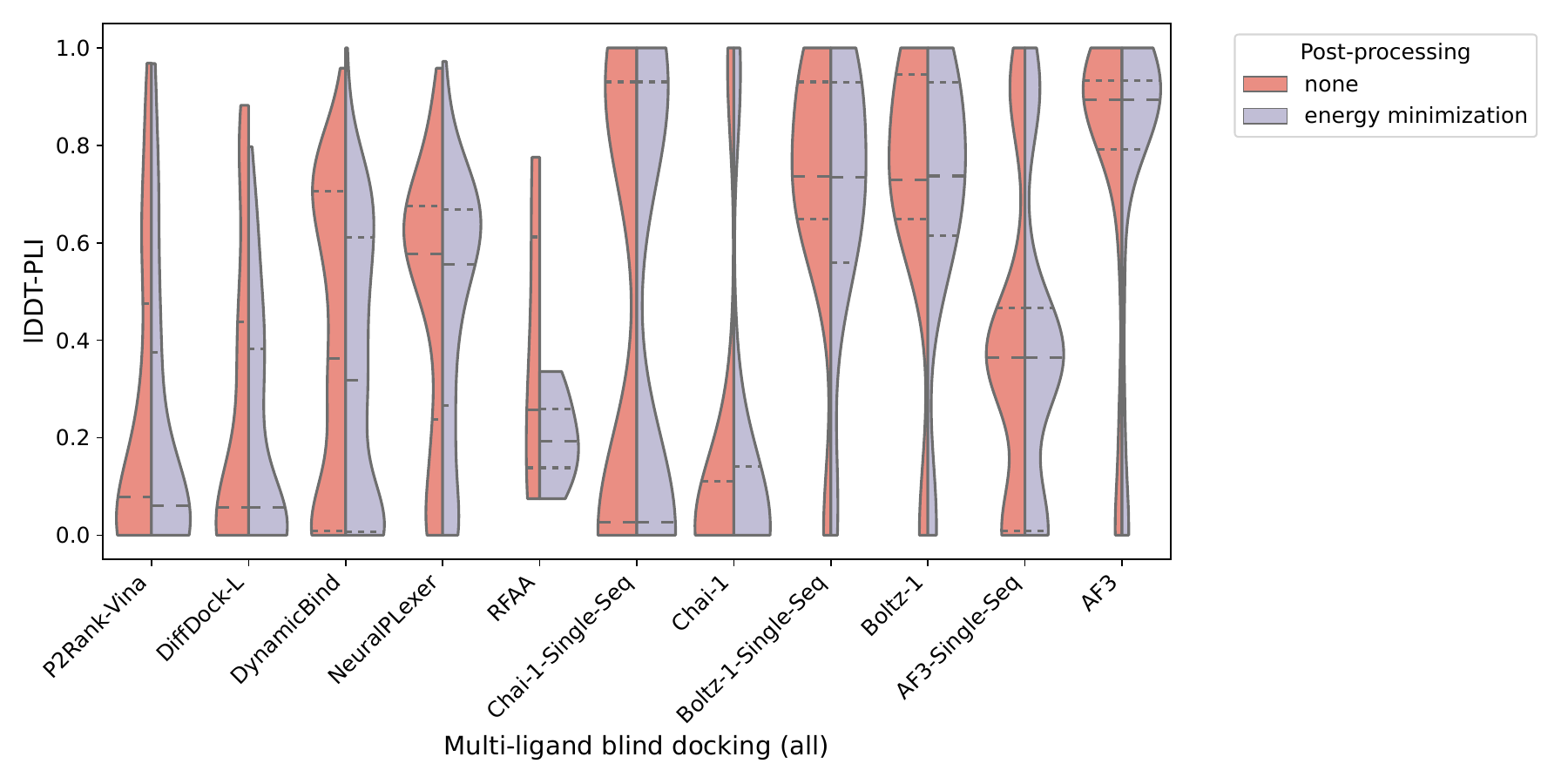}
  \caption{CASP15 dataset results for multi-ligand docking lDDT-PLI with relaxation.}
  \label{figure:casp15_multi_ligand_lddt_pli_results_with_relaxation}
\end{figure}

\begin{figure}
  \centering
  \includegraphics[width=\linewidth]{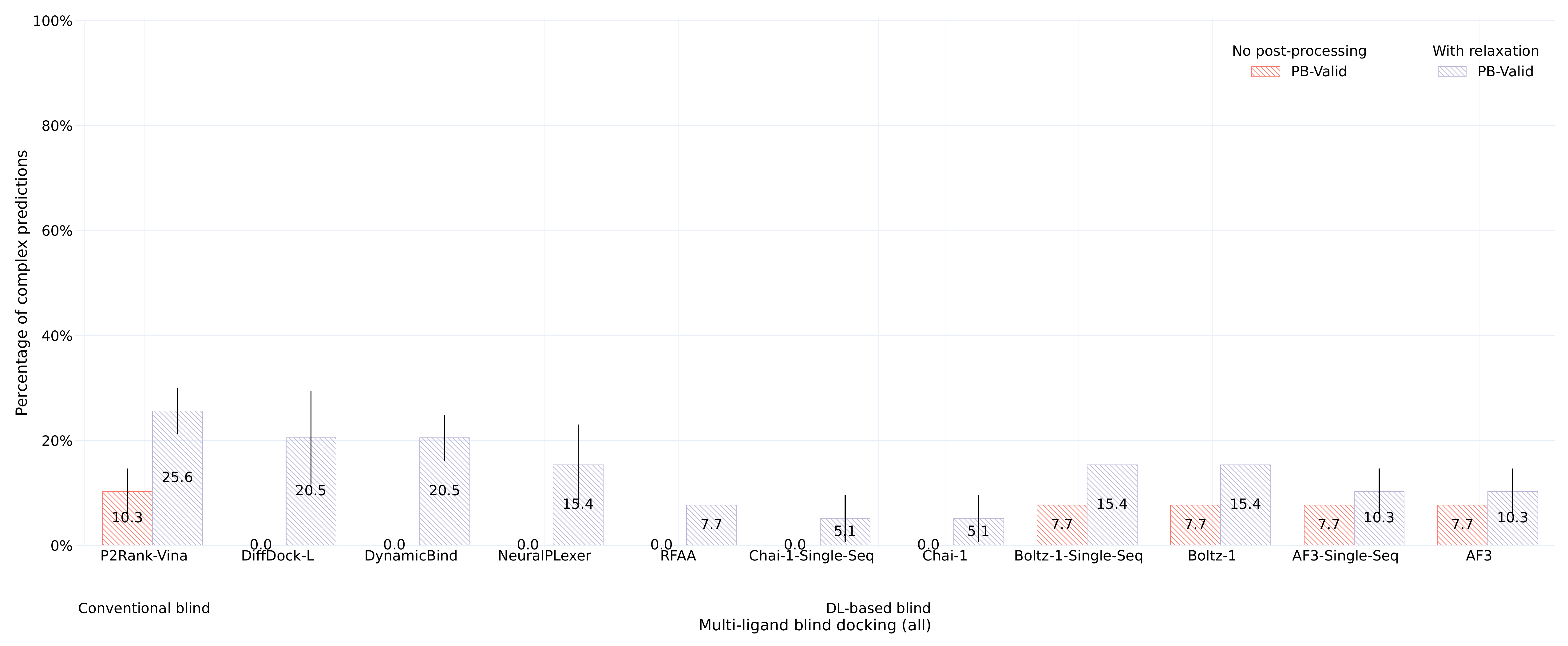}
  \caption{CASP15 dataset results for multi-ligand docking PB-Valid rates with relaxation. Data are presented as mean values +/- standard deviations over three independent predictions for each protein-ligand complex.}
  \label{figure:casp15_all_multi_ligand_relaxed_pb_valid_bar_chart}
\end{figure}

To start, Figures \ref{figure:casp15_multi_ligand_rmsd_results_with_relaxation}, \ref{figure:casp15_multi_ligand_lddt_pli_results_with_relaxation}, and \ref{figure:casp15_all_multi_ligand_relaxed_pb_valid_bar_chart} report each method's multi-ligand RMSD and lDDT-PLI distributions as well as PB-Valid rates with and without relaxation. We see that AF3 produces the most tightly bound and accurate RMSD and lDDT-PLI distributions overall yet is challenged in its PB-Valid rate by the conventional method AutoDock Vina, highlighting that AF3 predicted several structurally accurate yet chemically implausible multi-ligand conformations for this dataset.

\subsubsection{All single-ligand results}
\begin{figure}
  \centering
  \includegraphics[width=\linewidth]{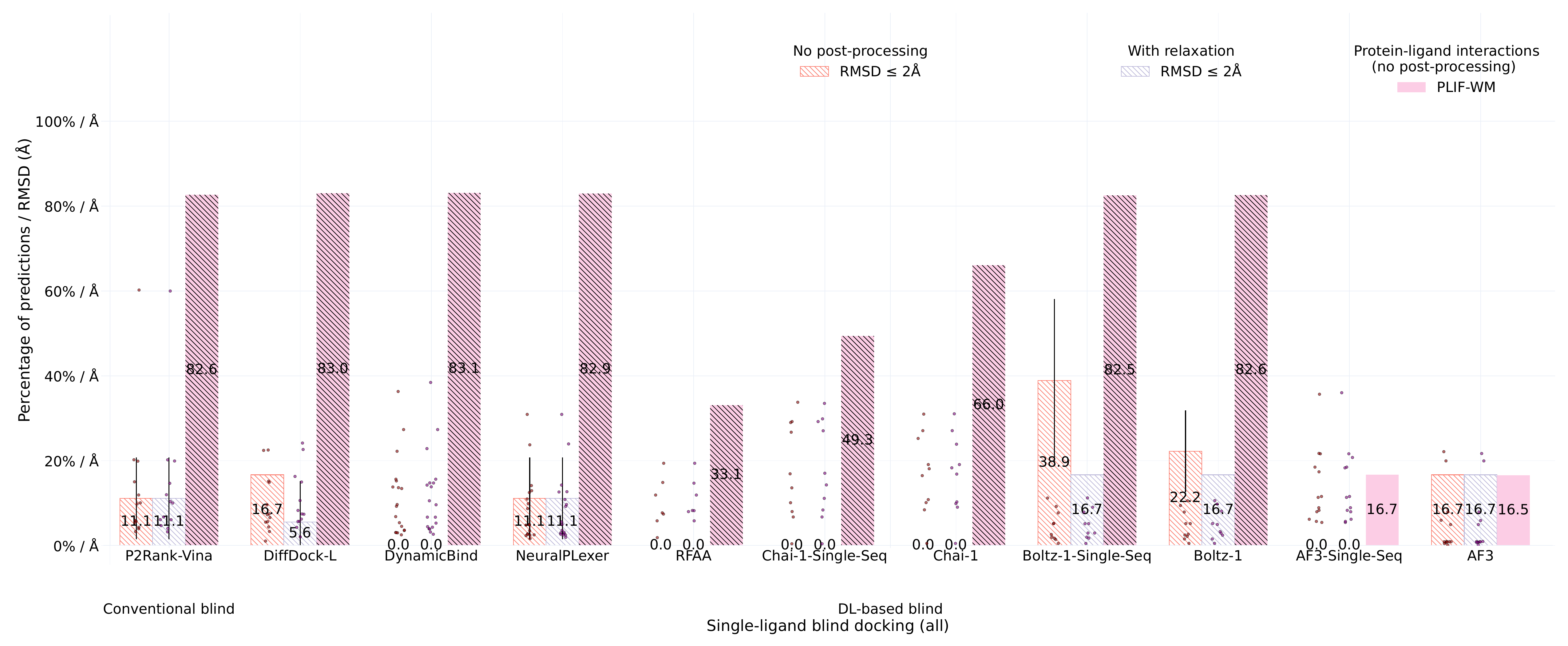}
  \caption{CASP15 dataset results for successful single-ligand docking with relaxation. Data are presented as mean values +/- standard deviations over three independent predictions for each protein-ligand complex.}
  \label{figure:casp15_single_ligand_docking_results_with_relaxation}
\end{figure}

\begin{figure}
  \centering
  \includegraphics[width=\linewidth]{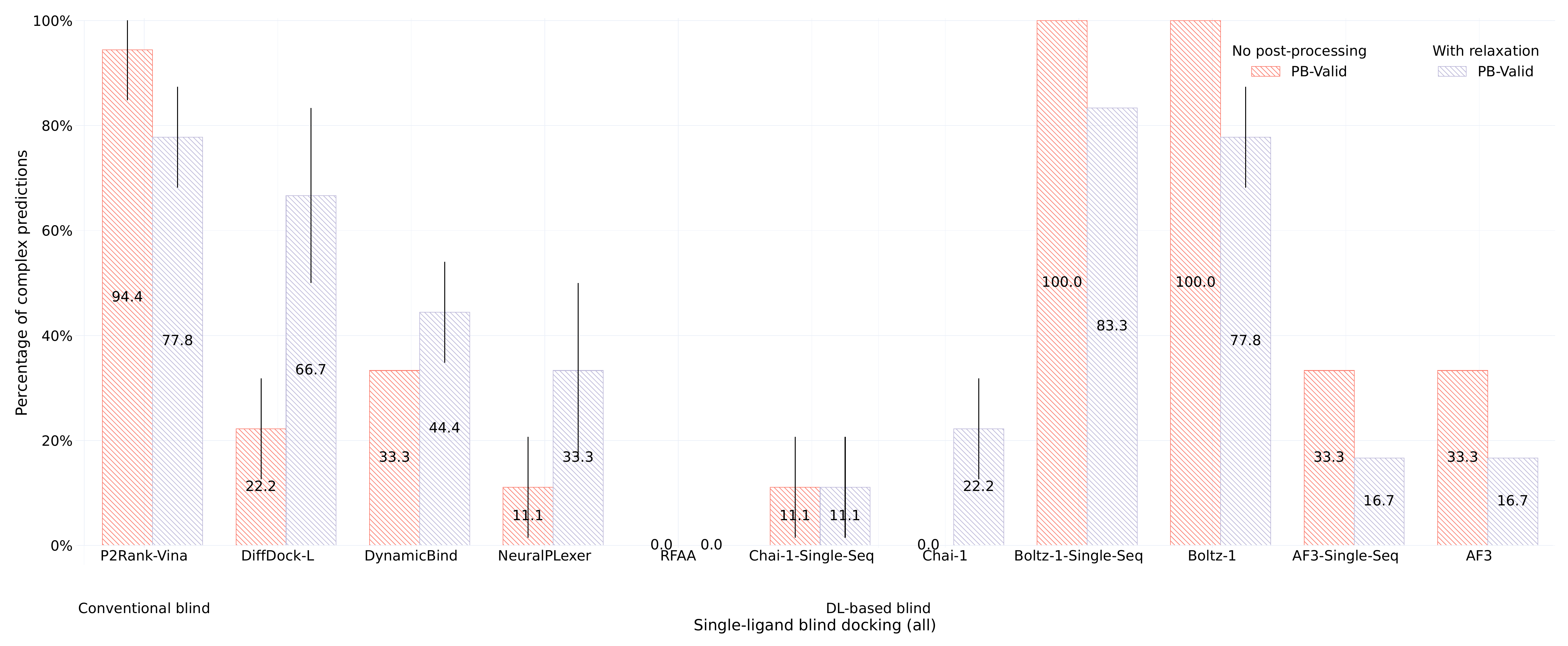}
  \caption{CASP15 dataset results for single-ligand PB-Valid rates with relaxation. Data are presented as mean values +/- standard deviations over three independent predictions for each protein-ligand complex.}
  \label{figure:casp15_single_ligand_pb_validity_results_with_relaxation}
\end{figure}

\begin{figure}
  \centering
  \includegraphics[width=\linewidth]{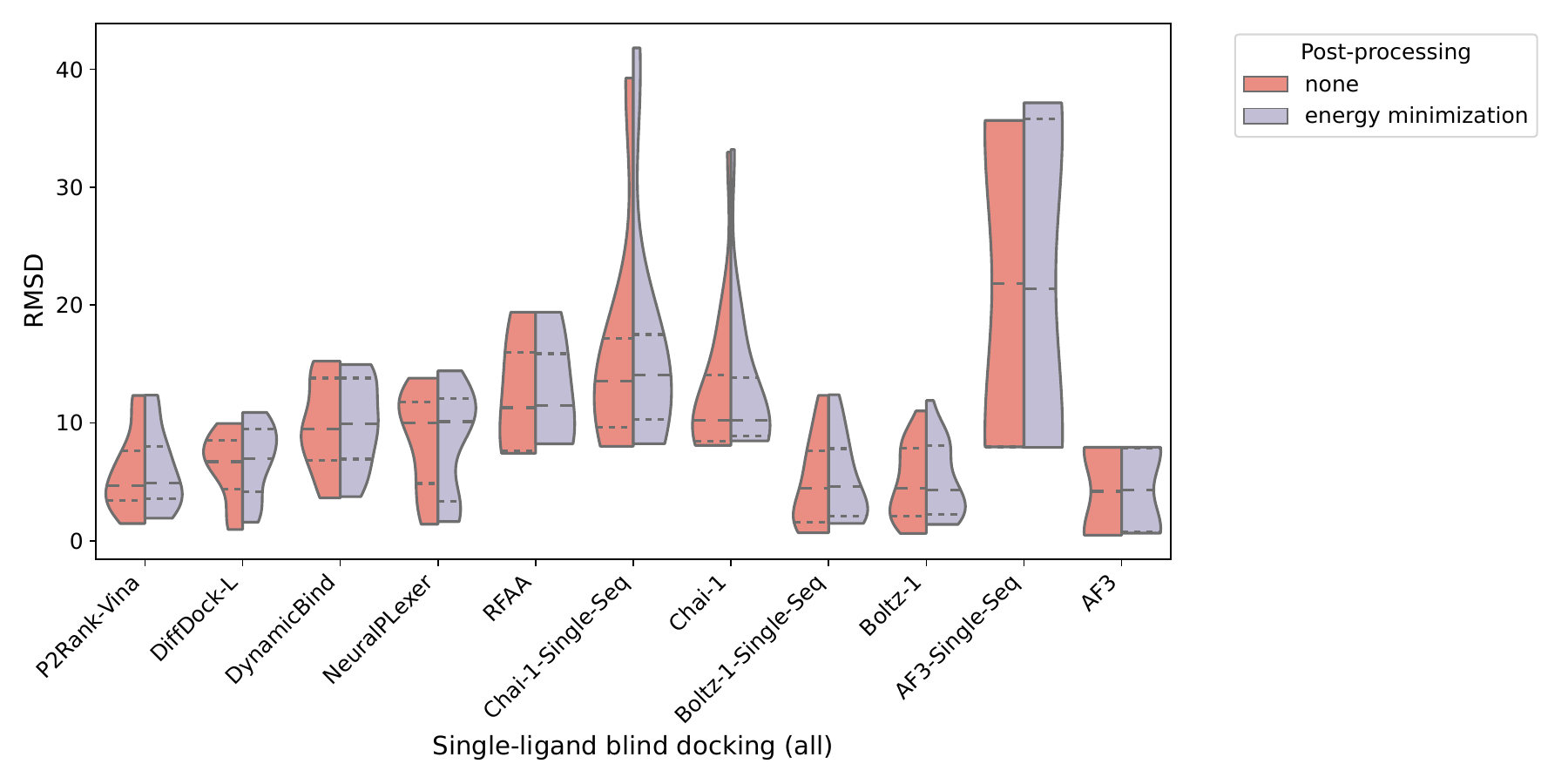}
  \caption{CASP15 dataset results for single-ligand docking RMSD with relaxation.}
  \label{figure:casp15_single_ligand_rmsd_results_with_relaxation}
\end{figure}

\begin{figure}
  \centering
  \includegraphics[width=\linewidth]{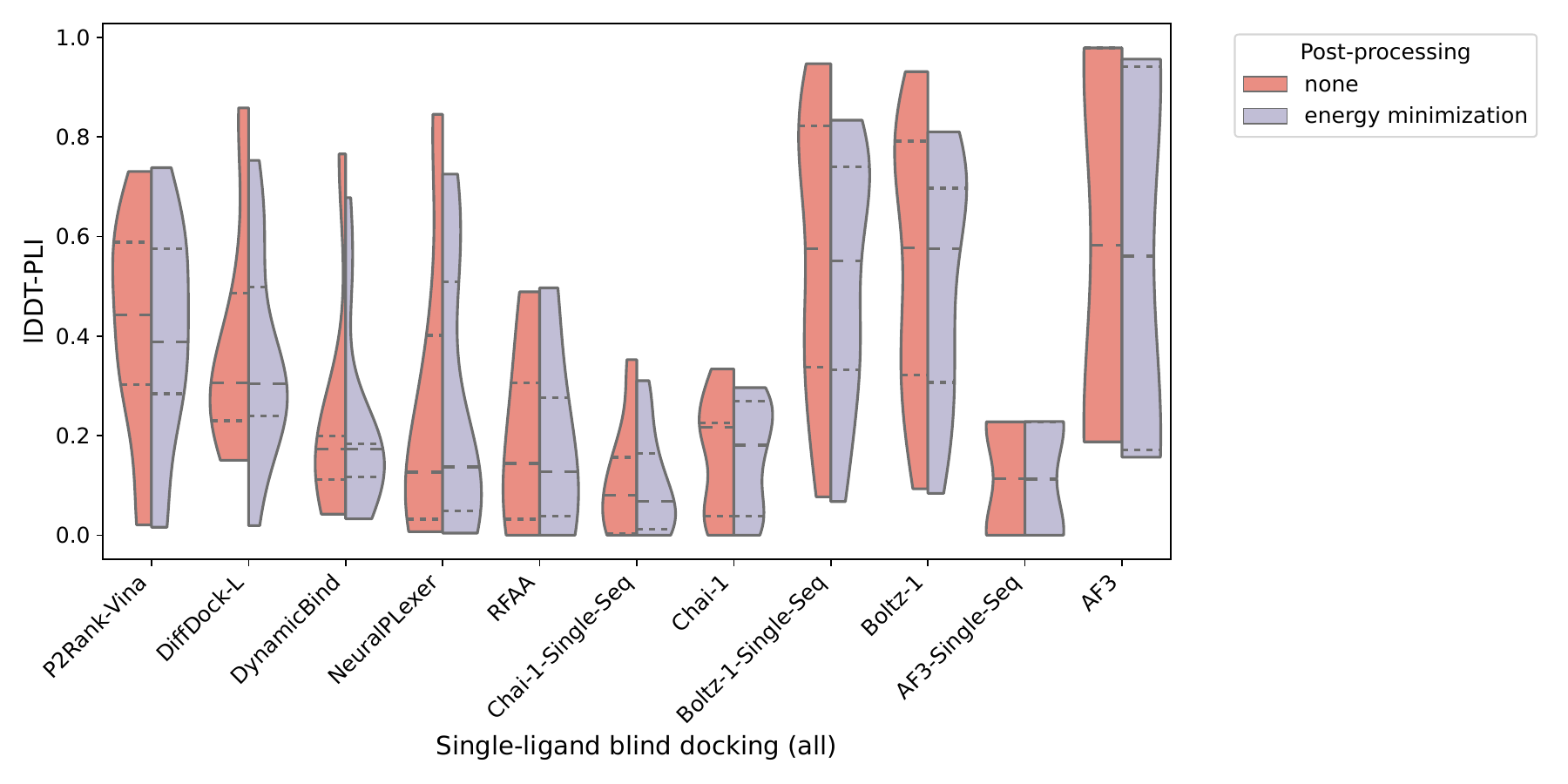}
  \caption{CASP15 dataset results for single-ligand docking lDDT-PLI with relaxation.}
  \label{figure:casp15_single_ligand_lddt_pli_results_with_relaxation}
\end{figure}

Next, Figures \ref{figure:casp15_single_ligand_docking_results_with_relaxation}, \ref{figure:casp15_single_ligand_pb_validity_results_with_relaxation}, \ref{figure:casp15_single_ligand_rmsd_results_with_relaxation}, and \ref{figure:casp15_single_ligand_lddt_pli_results_with_relaxation} display each method's single-ligand CASP15 docking success rates, PB-Valid rates, docking RMSD, and docking lDDT-PLI distributions, respectively. In summary, we can make a few respective observations from these plots. (1) Interestingly, Boltz-1 (with and without MSAs) achieves the highest structural and interaction modeling accuracy compared to AF3 and similar co-folding methods. (2) Even though most are positionally incorrect, structurally and chemically speaking, the majority of Boltz-1, AutoDock Vina, and DiffDock-L's predictions are valid according to the PoseBusters software suite, whereas few of AF3's predictions are. (3) Boltz-1 and AF3 yield notably lower RMSD distributions than all other baseline methods (including the similar DL co-folding method Chai-1). (4) Only AutoDock Vina, Boltz-1, and AF3 produce a reasonable range of lDDT-PLI scores for these single-ligand targets.

\subsubsection{Single and multi-ligand results for \textit{public} targets}
\begin{figure}
  \centering
  \includegraphics[width=\linewidth]{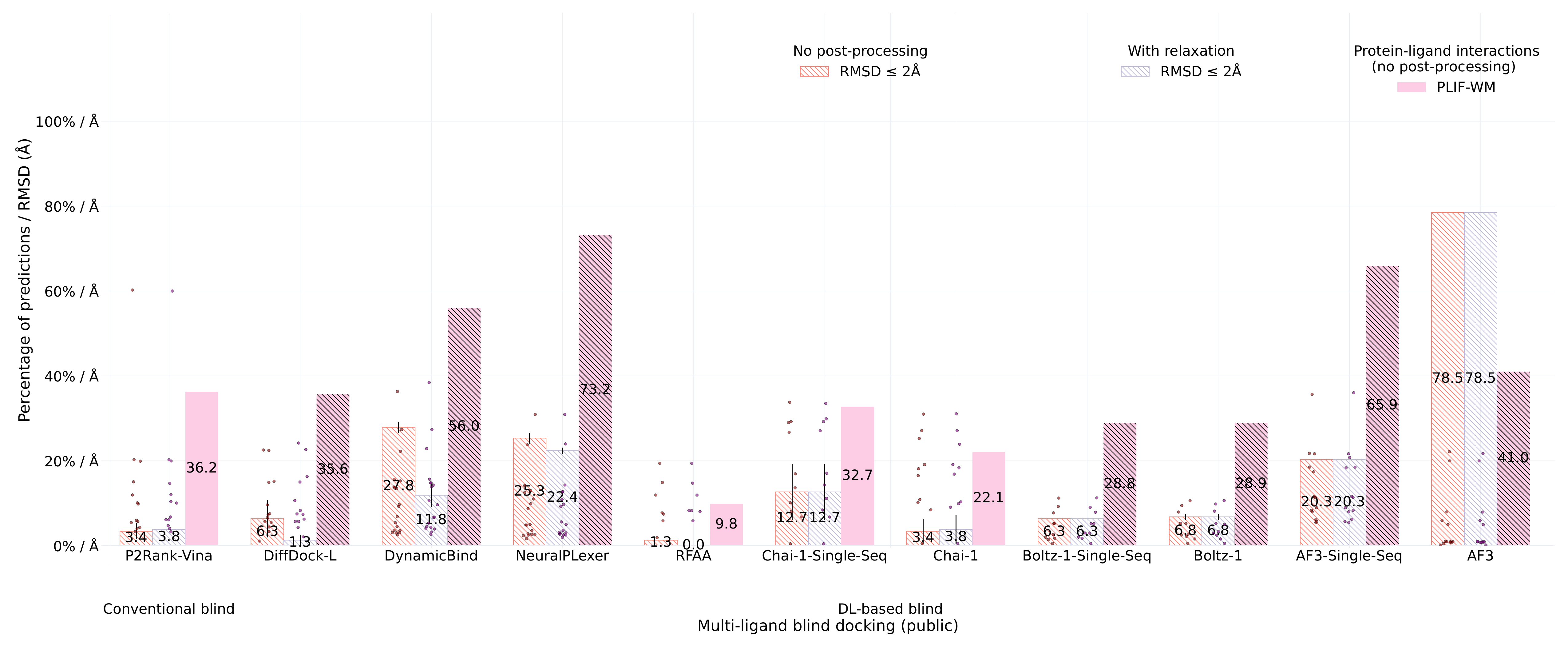}
  \caption{CASP15 public dataset results for successful multi-ligand docking with relaxation. Data are presented as mean values +/- standard deviations over three independent predictions for each protein-ligand complex.}
  \label{figure:casp15_public_multi_ligand_docking_results_with_relaxation}
\end{figure}

\begin{figure}
  \centering
  \includegraphics[width=\linewidth]{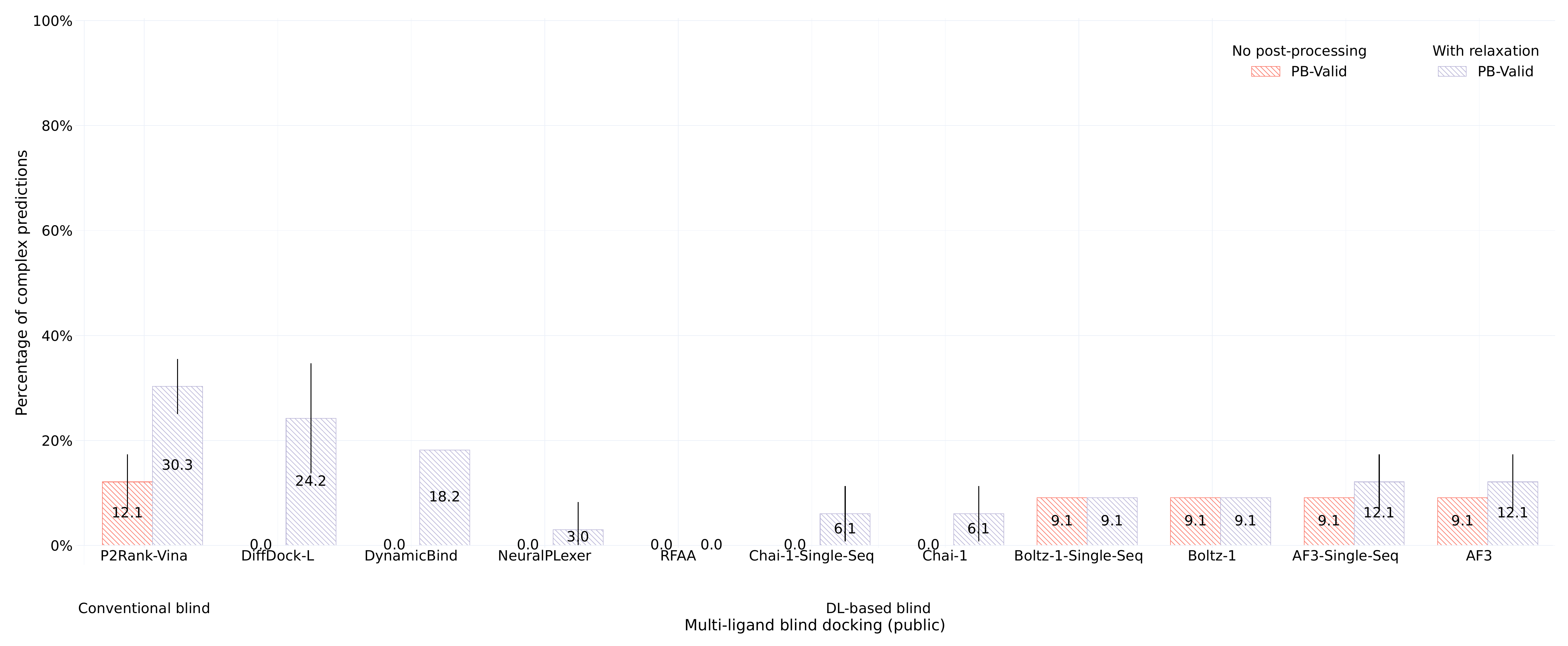}
  \caption{CASP15 public dataset results for multi-ligand PB-Valid rates with relaxation. Data are presented as mean values +/- standard deviations over three independent predictions for each protein-ligand complex.}
  \label{figure:casp15_public_multi_ligand_pb_validity_results_with_relaxation}
\end{figure}

\begin{figure}
  \centering
  \includegraphics[width=\linewidth]{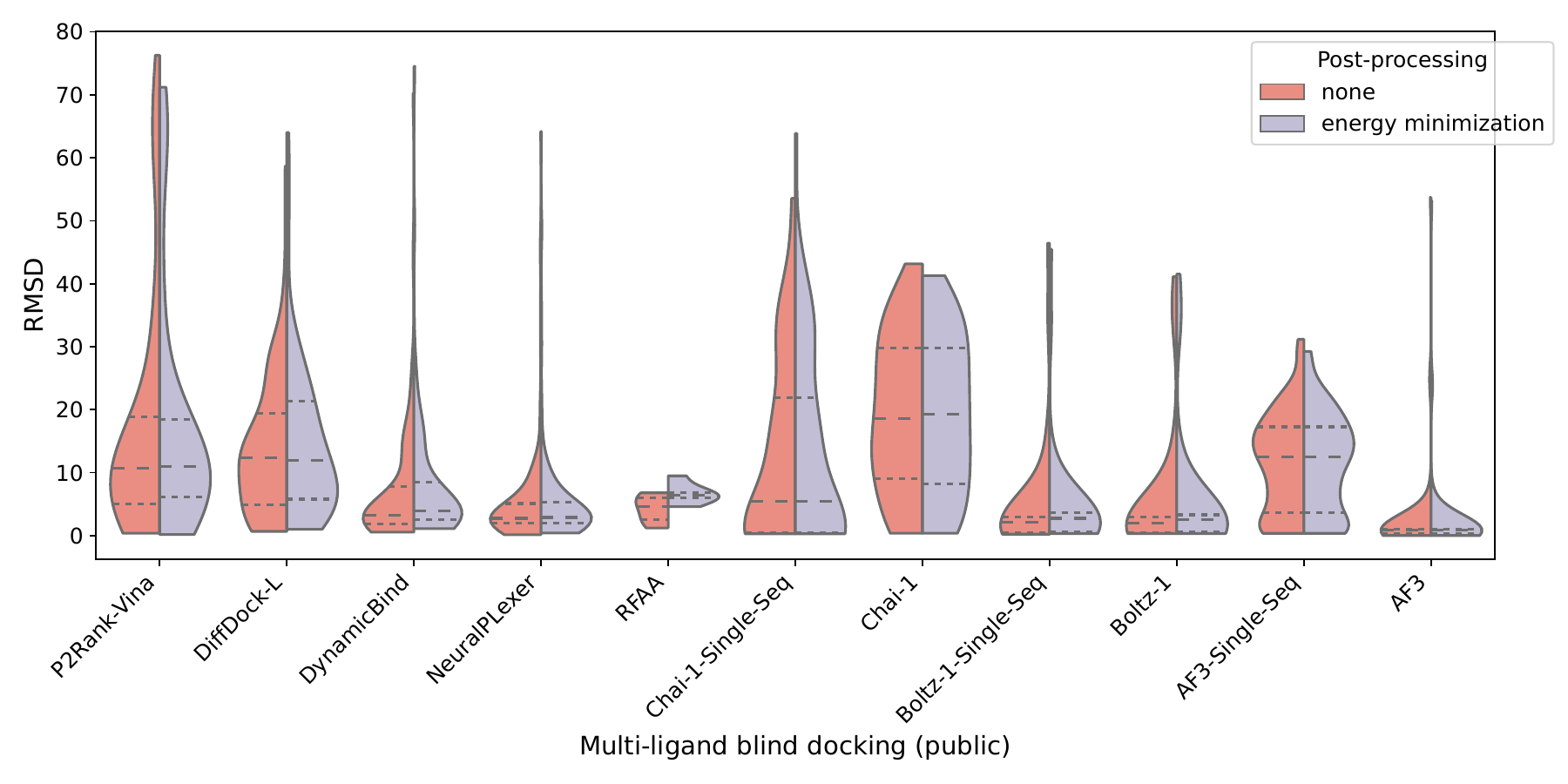}
  \caption{CASP15 public dataset results for multi-ligand docking RMSD with relaxation.}
  \label{figure:casp15_public_multi_ligand_rmsd_results_with_relaxation}
\end{figure}

\begin{figure}
  \centering
  \includegraphics[width=\linewidth]{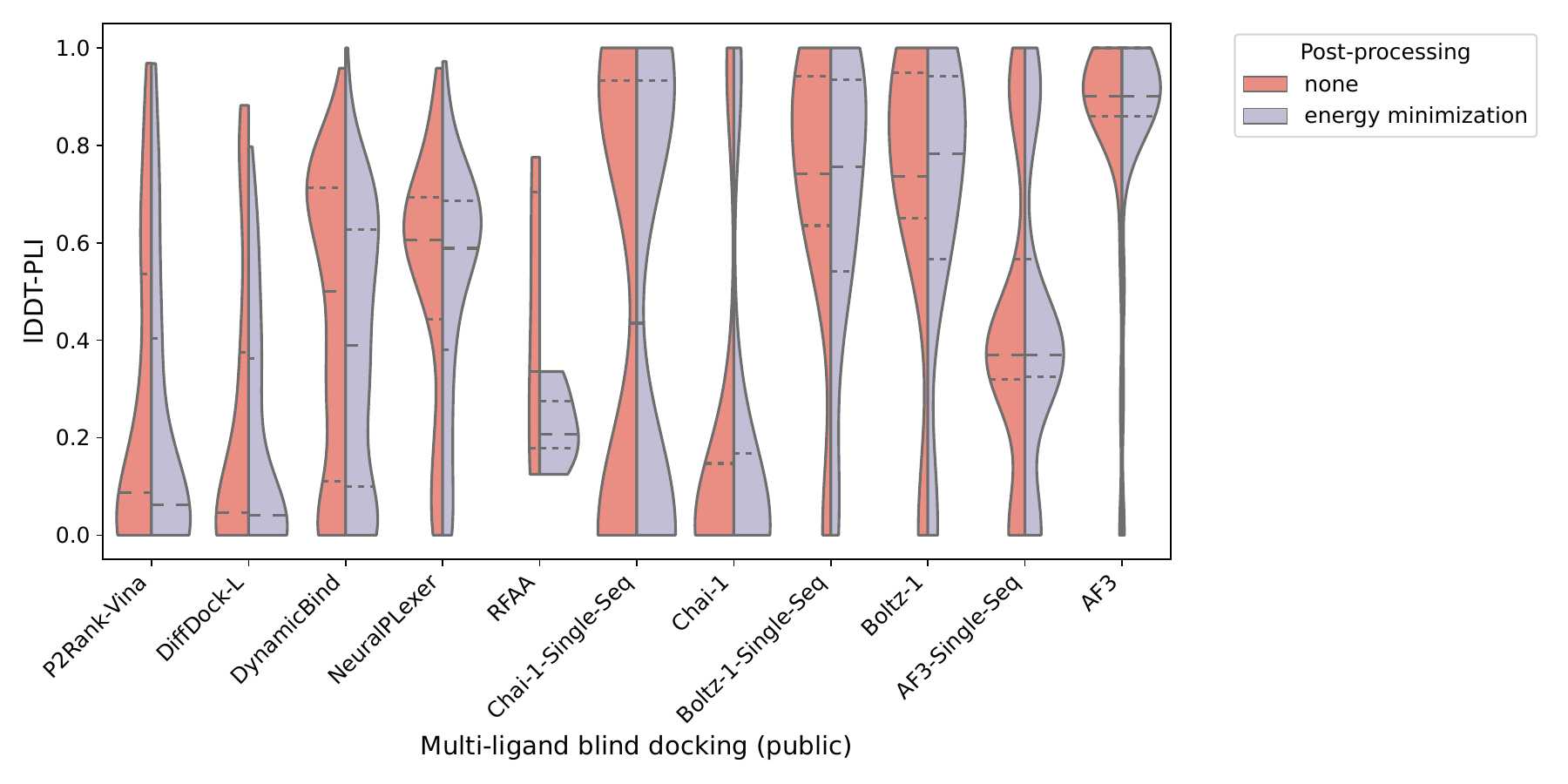}
  \caption{CASP15 public dataset results for multi-ligand docking lDDT-PLI with relaxation.}
  \label{figure:casp15_public_multi_ligand_lddt_pli_results_with_relaxation}
\end{figure}

\begin{figure}
  \centering
  \includegraphics[width=\linewidth]{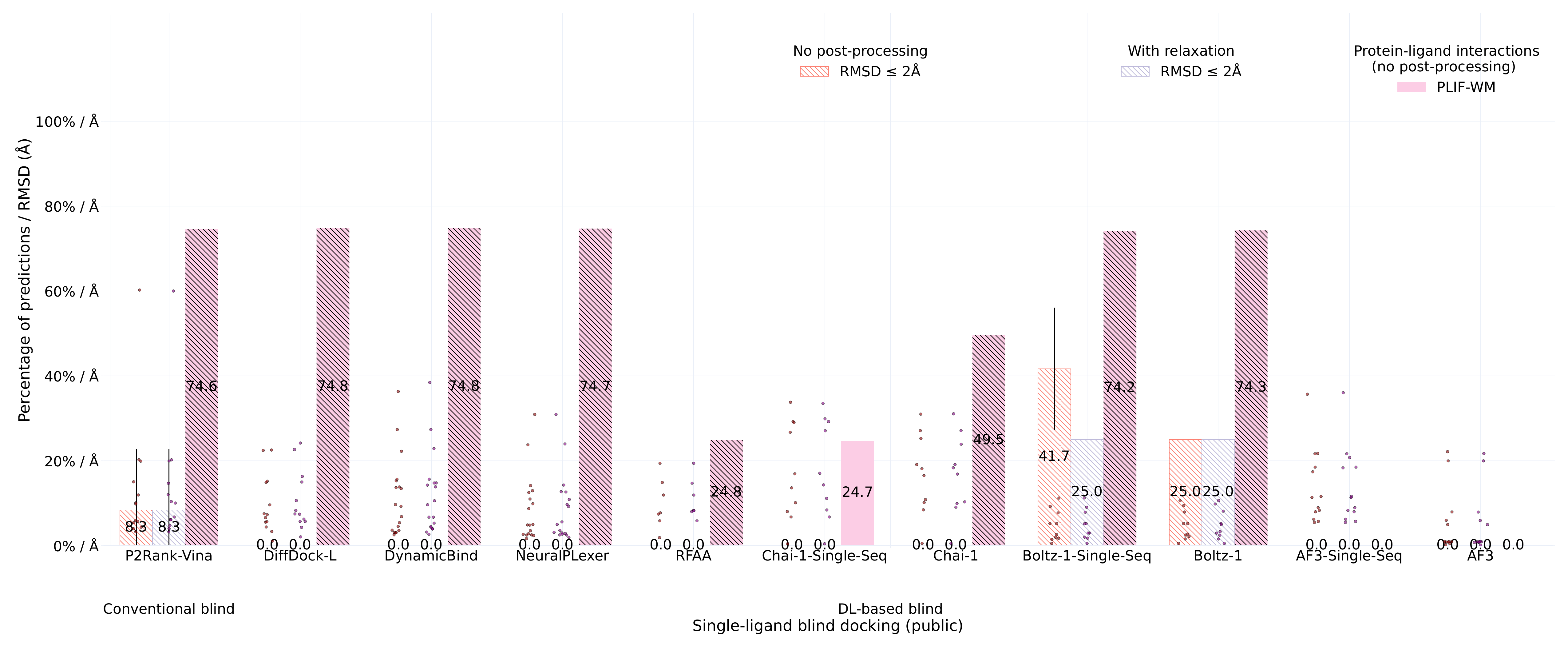}
  \caption{CASP15 public dataset results for successful single-ligand docking with relaxation. Data are presented as mean values +/- standard deviations over three independent predictions for each protein-ligand complex.}
  \label{figure:casp15_public_single_ligand_docking_results_with_relaxation}
\end{figure}

\begin{figure}
  \centering
  \includegraphics[width=\linewidth]{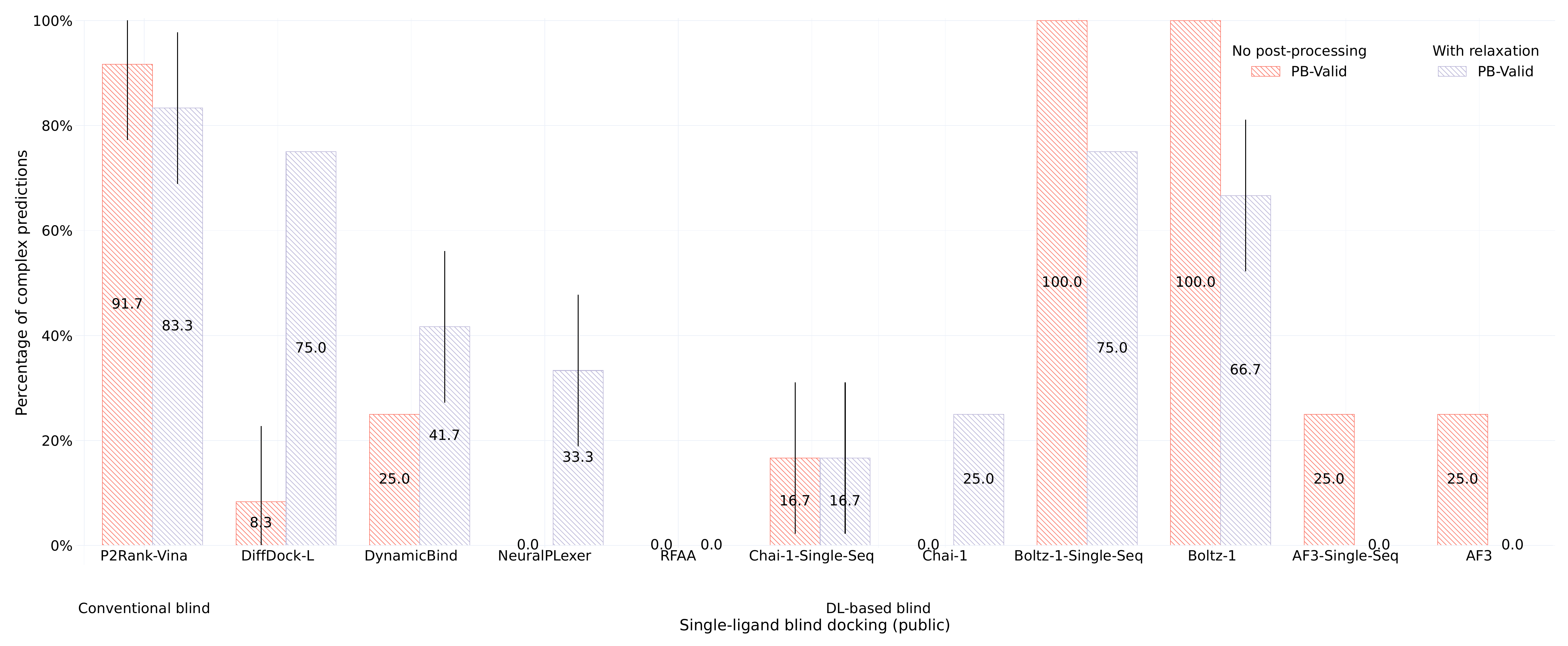}
  \caption{CASP15 public dataset results for single-ligand PB-Valid rates with relaxation. Data are presented as mean values +/- standard deviations over three independent predictions for each protein-ligand complex.}
  \label{figure:casp15_public_single_ligand_pb_validity_results_with_relaxation}
\end{figure}

\begin{figure}
  \centering
  \includegraphics[width=\linewidth]{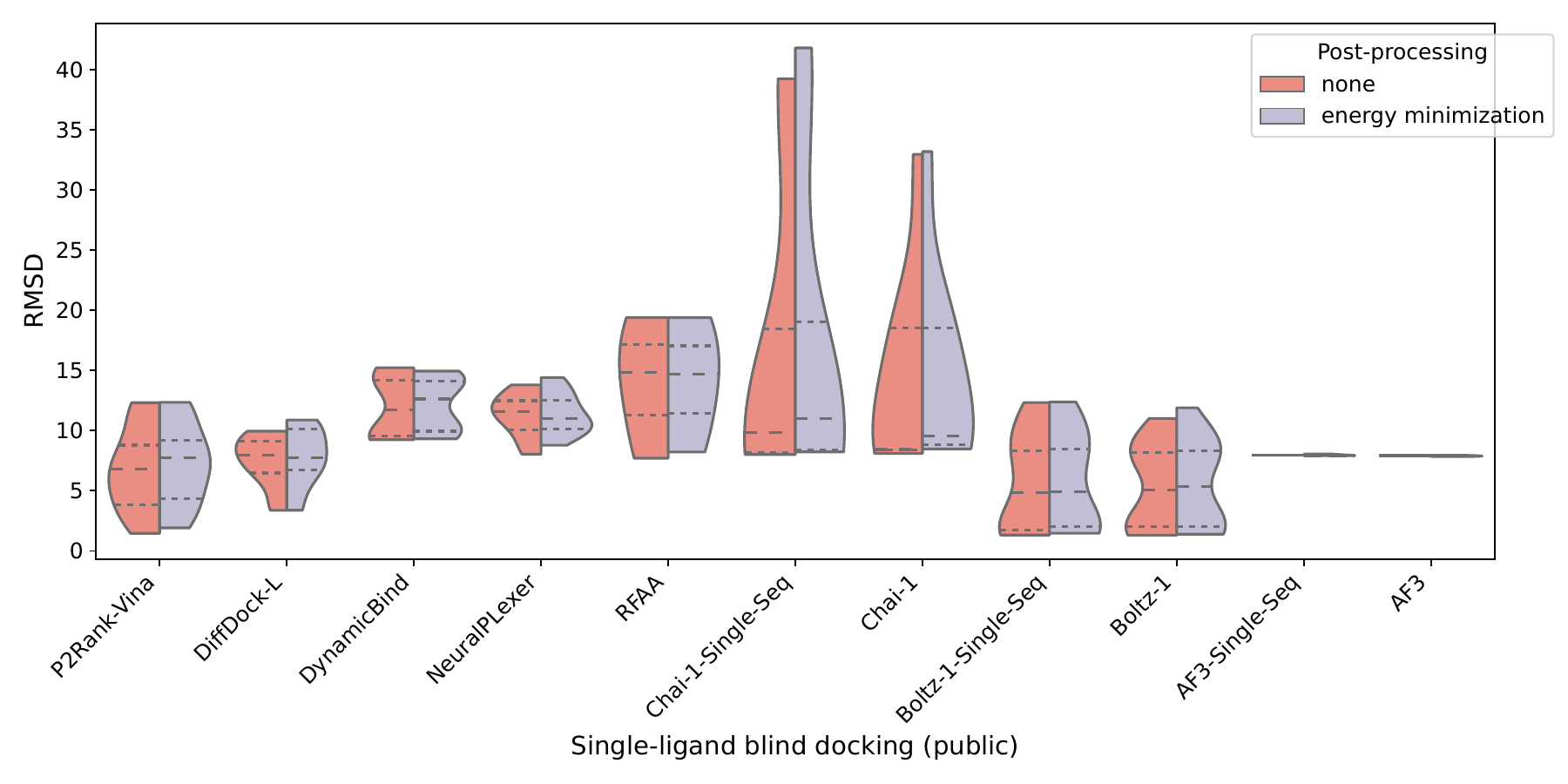}
  \caption{CASP15 public dataset results for single-ligand docking RMSD with relaxation.}
  \label{figure:casp15_public_single_ligand_rmsd_results_with_relaxation}
\end{figure}

\begin{figure}
  \centering
  \includegraphics[width=\linewidth]{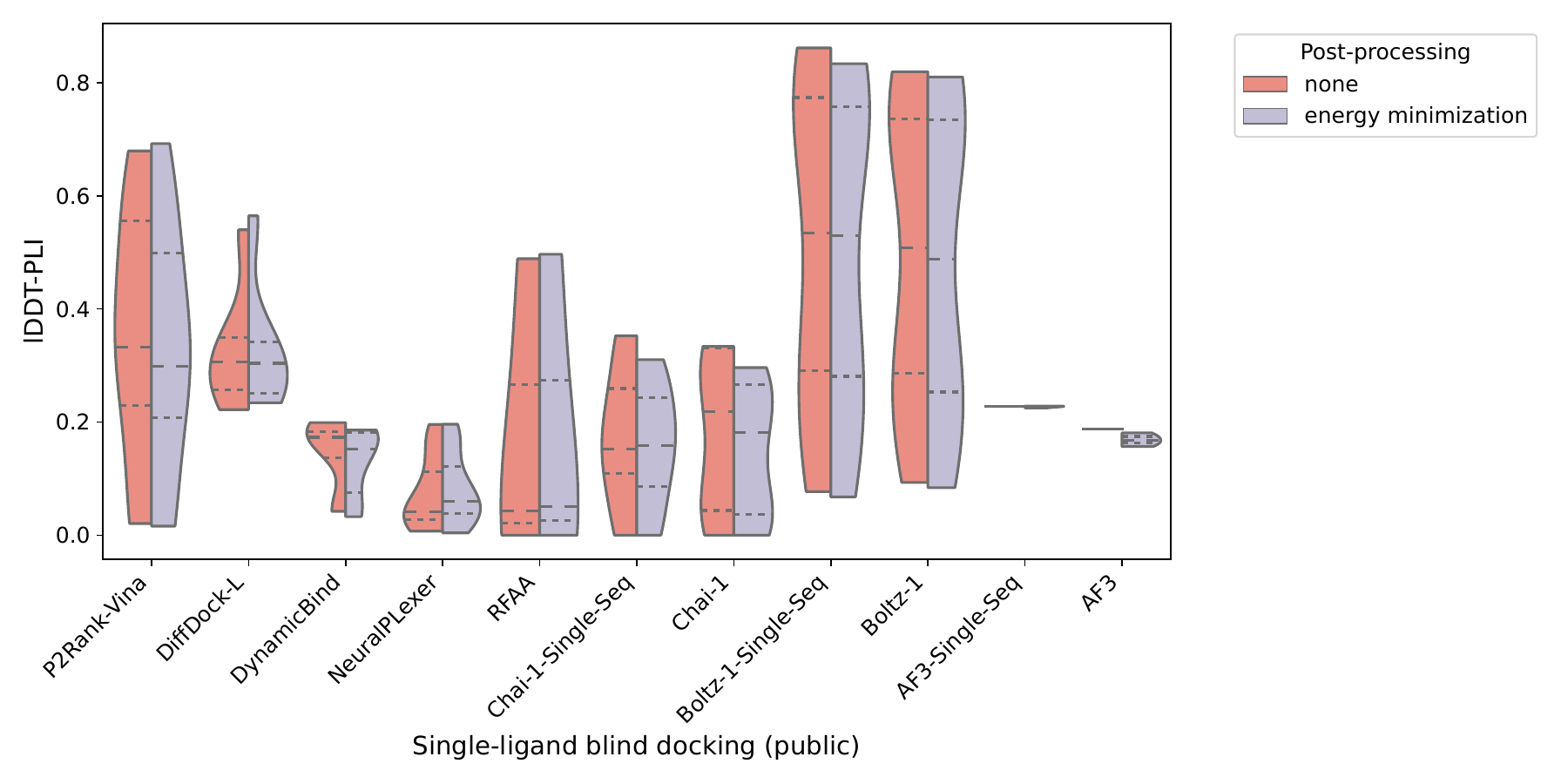}
  \caption{CASP15 public dataset results for single-ligand docking lDDT-PLI with relaxation.}
  \label{figure:casp15_public_single_ligand_lddt_pli_results_with_relaxation}
\end{figure}

Lastly, for completeness and reproducibility, Figures \ref{figure:casp15_public_multi_ligand_docking_results_with_relaxation}, \ref{figure:casp15_public_multi_ligand_pb_validity_results_with_relaxation}, \ref{figure:casp15_public_multi_ligand_rmsd_results_with_relaxation}, and \ref{figure:casp15_public_multi_ligand_lddt_pli_results_with_relaxation} present corresponding multi-ligand results for the public CASP15 targets, whereas Figures \ref{figure:casp15_public_single_ligand_docking_results_with_relaxation}, \ref{figure:casp15_public_single_ligand_pb_validity_results_with_relaxation}, \ref{figure:casp15_public_single_ligand_rmsd_results_with_relaxation}, and \ref{figure:casp15_public_single_ligand_lddt_pli_results_with_relaxation} report corresponding single-ligand results for the public CASP15 targets. Overall, we observe marginal differences between the full and public CASP15 target results for multi-ligand complexes, since once again AF3 achieves top results overall in the context of multi-ligands. However, for most methods, we notice more striking performance drops between the full and public \textit{single}-ligand CASP15 target results, suggesting that some of the private single-ligand complexes are easier prediction targets than most of the publicly available single-ligand complexes. In short, we find that Boltz-1 consistently performs best in this single-ligand setting.

\clearpage

\clearpage

\clearpage
\clearpage

\clearpage

\clearpage
\clearpage

\clearpage
\clearpage

\clearpage
\clearpage

\end{appendices}




\end{document}